**Social Learning and Diffusion of Pervasive Goods:**

**An Empirical Study of an African App Store**

Meisam Hejazi Nia

Brian T. Ratchford

Norris Bruce

*** Please do not cite, quote, or distribute without permission ***

Meisam Hejazi Nia is a senior data scientist at Staples (meisam.hejazynia@gmail.com), Brian T. Ratchford (972-883-5975, btr051000@utdallas.edu) is Charles and Nancy Davidson Professor in Marketing, and Norris I. Bruce (972.883.6293, nxb018100@utdallas.edu) is an Associate Professor at the Naveen Jindal School of Management at the University of Texas at Dallas, P.O. Box 830688, Richardson, TX 75083.

**Social Learning and Diffusion of Pervasive Goods:**

**An Empirical Study of an African App Store**

*ABSTRACT*

In this study, the authors develop a structural model that combines a macro diffusion model with a micro choice model to control for the effect of social influence on the mobile app choices of customers over app stores. Social influence refers to the density of adopters within the proximity of other customers. Using a large data set from an African app store and Bayesian estimation methods, the authors quantify the effect of social influence and investigate the impact of ignoring this process in estimating customer choices. The findings show that customer choices in the app store are explained better by offline than online density of adopters and that ignoring social influence in estimations results in biased estimates. Furthermore, the findings show that the mobile app adoption process is similar to adoption of music CDs, among all other classic economy goods. A counterfactual analysis shows that the app store can increase its revenue by 13.6% through a viral marketing policy (e.g., a sharing with friends and family button).

Smartphones pervade the global telecommunications market to such an extent that in the United States, for example, a consumer can adopt a smartphone handset with a postpaid contract, no matter which mobile operator (e.g., T-Mobile, Verizon, AT&T) he or she selects. The smartphone handsets and mobile apps are complements. A mobile app store (e.g., Google play, Apple and Microsoft app stores) acts as a two-sided platform that matches consumers to mobile app publishers/developers. The mobile app platform revenue comes from two sources: selling the paid apps or advertising on freemium apps. As a result, for the app store platform, consumer adoption of mobile apps represents a critical problem. The app store platform has a large amount of information about consumers' download behavior, enabling it to customize its marketing actions to target different consumers according to their behaviors. For example, a mobile app platform can decide to offer viral-referral or free-trial strategies. A viral-referral strategy is useful when consumers' preferences are interrelated because of the psychological benefits of social identification, learning, and inclusion and the utilitarian benefits of network externalities. A free-trial strategy is useful when consumers have learning costs or are uncertain about a mobile app.

It is common for customers to have interrelated preferences for mobile apps. Online forums are filled with questions about requests for mobile app recommendations,[1] and app stores try to inform users about the popularity of mobile apps. The interdependence of mobile app choices is important because customers often do not know what mobile app they want, so they rely on offline family, friends, and colleagues to find new apps. App stores have tried to facilitate this process by creating "Tell a Friend" and "Share This Application" (WonderHowTo 2011). Therefore, to design policies to influence consumers' mobile app choices, an app store platform needs a framework to quantify not only the effect of mobile app characteristics but also the effect of online and offline social influences on customer choices.

---

[1] See "Title of actual blog here" at http://www.cnet.com/forums/mobile-apps/.

Given this context, we ask the following questions:

1. How can we design a targeting approach for an app store platform?

2. How does the social learning process of mobile app customers differ from that of classic economy goods, such as a color television?

3. How can an app store platform capture the heterogeneity of its customers and the variation in mobile apps to customize its marketing actions?

4. What are the key elements of consumers' utility of adopting a mobile app that allows an app store platform to group and target potential customers?

To answer these questions, we combine a macro social learning diffusion model with a micro choice model and use the choice model to examine consumers' adoption behavior. To control for social influence, we apply a filtering technique to another aggregated data set to create a time-varying measure of social influence. Also to control for mobile app characteristics, price, and advertising, we use a factor model. We run the filtering technique on two aggregate adoption data sets for approximately 200 days. These data sets include, on the one hand, the cumulative number of adopters in a local city in Africa and, on the other hand, the cumulative number of adopters across all 30 cities in which the platform of interest globally operates. We refer to these two data sets as the aggregate data sets hereinafter. We run the choice model and the factor model on a data set of a sample of choices of 147 consumers over 20 weeks; we refer to this sample as the micro sample.

We use the social learning diffusion model of Van den Bulte and Joshi (2007) to model the simultaneous diffusion of mobile apps on the app store. Such a modeling approach presents two challenges. The first concerns mobile app consumers' choice sparse data, because the download of a mobile app is a rare event. To address this challenge, we aggregate the data at an app-category level. The second challenge involves dealing with the possible measurement error. For this purpose, we cast Van den Bulte and Joshi's model into a discrete-time state-space model. The use of the Gaussian process to filter the measurement error is also quite common in online mission critical systems, such as robotics. In this case, we filter two double-degree polynomial differential equations of each mobile app category's diffusion. We use an unscented Kalman filter (UKF), an approach introduced to machine learning and robotics to estimate the nonlinear diffusion equation up to third-order precision (Julier and Uhlmann 1997; Wan and Van Der Merwe 2001). This approach is an alternative to the extended Kalman filter, which estimates the nonlinear diffusion equation only up to first-order precision.

We also employ a hierarchical prior with a seemingly unrelated regression (SUR) model to use the shared information in the simultaneous diffusion of the mobile app categories and to avoid overfitting the model with 300 macro parameters. To estimate the macro diffusion model in the short planning time horizon of an app store platform, we use a Monte Carlo expectation maximization (MCEM) approach to optimize the maximum a posteriori (MAP) of the parameters, in contrast with possibly slow convergence Bayesian sampling algorithms, such as Gibbs and Metropolis–Hastings. To deal with the problem of the stochastic surface search of the MCEM approach, we use a genetic optimization algorithm, with an initial population that is a perturbed version of the estimates found in Van den Bulte and Joshi's (2007) study.

Next, we use the outcome of the macro diffusion model as a measure for social influence in the structural choice model to extract factors of customers' mobile app choices. The choice of a mobile app adoption is quite sparse over time. In other words, we expect to observe several zeros in the data. To deal with such sparsity and filter out the possible noise of the data, we aggregate the data on the characteristics of the mobile app categories and the cumulative number of imitators at a weekly level. Furthermore, so as not to discard the multicollinear data on the mobile app characteristics, we use a factor model to recover the underlying factors of the mobile app profiles. To name these factors, we merge the factor loading profiles and practitioners' knowledge of customers' mobile app choices.

Given the mobile app category latent factors, the density of the imitators, and the download history of the app store platforms, we use a mixture normal multinomial logit model to represent each consumer's choice of mobile app adoption. We estimate this model with the Markov chain Monte Carlo (MCMC) sampler. The hierarchical modeling and the weighting scheme used make the approach appropriate for the big data, because the mixture normal prior allows for a flexible structure that may not overfit. This modeling approach is appropriate for the context of online retailers, in which the distribution of choices follows a long-tail distribution (Anderson 2006).

We estimate this model over a data set of a newly launched app store in Africa during May 2013 and a supplementary data set of network location of the mobile app store users obtained from the web. The sample consists of mobile app choices of approximately 20,000 customers residing in 30 cities in which the app store is available, among which approximately 3000 customers reside in a city in Africa. Mobile apps belong to various categories, among which we selected ten categories that were less sparse. The estimation results show that social influence significantly affects customer app adoption choices and that offline social influences (in the local city) explain customers' choices better than online social influences (in the 30 cities in which the app store is located). We also find that not controlling for social influence in consumers' mobile app choices results in biased customer preference estimates. Furthermore, we find that among different classic economy goods, the mobile app adoption pattern is similar to that of music CDs.

We also use the estimated micro choice model to simulate a counterfactual policy that intervenes in social influence to affect consumers' choices. We find a policy that increases mobile app diffusion by 13.6%. This step is a form of prescriptive analytics built from the descriptive and predictive analytics steps. Furthermore, we find individual-specific preference parameters estimated by the choice model, which can help the mobile app store target its customers.

This study is related not only to research on consumers' peer effect (e.g., Nair, Manchanda, and Bhatia 2010; Stephen and Toubia 2010; Yang and Allenby 2003) but also to research on global macro diffusion (e.g., Dekimpe, Parker, and Sarvary 1998; Putsis et al. 1997; Van den Bulte and Joshi 2007). Another relevant research stream includes studies on micro diffusion models (e.g., Chaterjee and Eliashberg 1990; Dover, Goldenberg, and Shapira 2012; Young 2009). The last stream of relevant studies includes studies on the app store platform (e.g., Carare 2012; Garg and Telang 2013; Ghose, Goldfarb, and Han 2012; Ghose and Han 2011a, b, 2014; Kim et al. 2010; Liu, Au, and Choi 2012). Although these studies have contributed greatly to the understanding of the phenomenon, none has created a pipeline that combines the macro diffusion modeling and the micro structural choice modeling approaches to allow the app store platform to target its consumers. In contrast, the proposed approach allows this by applying descriptive, predictive, and prescriptive analytics over high-volume, high-velocity, high-variety, and high-veracity big data.

Thus, this paper contributes to the emerging literature on prescriptive data analytics of the mobile app store platform in five ways. First, it introduces the combination of a macro simultaneous social learning adoption model and a micro structural choice modeling approach to design a method that allows app store platforms to target their heterogeneous consumers, using big data. Second, this paper benchmarks the parameters of social learning mobile app adoption against those of classic economy goods, such as television, personal computers, music CDs, and radio head. Third, this paper shows that offline social influence (at the local city level) drives customers' mobile app choices and that ignoring social learning processes creates biased estimates. Fourth, this paper shows the power of the proposed model for prescriptive analytics over big data, by finding an optimal viral marketing policy (e.g., share with friends and family) for the app store that can increase its total expected diffusion by 13.6%. Fifth, to estimate the proposed social learning model, this paper employs UKF, SUR, MCEM, and a genetic algorithm to maximize the MAP estimate of the macro diffusion model. In addition, it uses a hierarchical mixture normal prior over the multinomial logit choice model, estimating the model with the MCMC sampling method. These approaches, which allow for a flexible heterogeneity pattern and a robust filtering of process and measurement errors, as well as computational feasibility of big data analytics, should be of interest to academia and commercial entities concerned about the descriptive, predictive, and prescriptive analytics of their big data.

*LITERATURE REVIEW*

This study draws on several research streams investigated in the literature, including (1) the interdependence of consumer preference, (2) mobile app store dynamics, and (3) global macro and micro diffusion and social learning. Given the breadth of these areas across multiple disciplines, the following discussion represents only a brief review of these relevant streams. Table 1 presents a summary of the position of this study in the literature.

--- Insert Table 1 here--

*Interdependence of Consumer Preference*

Quantitative models of consumer purchase behavior often do not recognize that consumers' choices may be driven by the underlying social learning processes within the population. Economic models of choice typically assume that an individual's latent utility is a function of the brand and attribute preferences, rather than the preferences of the other customers. However, for pervasive experience goods, a new model that accounts for these underlying forces and preferences may better explain consumers' choices. Many studies have tried to address this issue, using cross-sectional data to model consumers' preference dependency (Yang and Allenby 2003), online social network seller interaction data to quantify the network value of consumers (Stephen and Toubia 2010), customer trial data to determine the importance of consumers' spatial exposure (Bell and Song 2007), and physicians' prescription choices and self-reported information to demonstrate the significant effect of network influence on consumers' choices (Nair, Manchanda, and Bhatia 2010).

Other researchers have also reported on the critical role of social proximity in shaping consumer preferences. Bradlow et al. (2005) build on previous literature to suggest that demographic and psychometric proximity measures are important for consumers' choice. Hartmann (2010) uses customer data to show a correlation between social interactions and the equilibrium outcome of an empirical discrete game. Yang, Narayan, and Assael (2006) demonstrate the interdependence of spouses' television viewership to suggest the need for considering choice interdependency. Narayan, Rao, and Saunders (2011) employ conjoint experience data to highlight the effects of peer influences. Finally, Choi, Hui, and Bell (2010) draw from an Internet retailer's data set to establish the importance of imitation effects in terms of geographical and demographical proximity. However, although these studies are significant in suggesting the role of social influence in decision making, none has modeled the interdependence of consumers' mobile app choices.

*Mobile App Store Dynamics*

Recently, a stream of literature has emerged that pertains to the dynamics of mobile app stores. Some studies have addressed Apple and the Google platforms' competition (Ghose and Han 2014), Google Play's freemium strategy (Liu, Au, and Choi 2012), and the impact of Apple's app store's best-seller rank information on sales (Carare 2012; Garg and Telang 2013). Other studies have addressed the relationship between content generation and consumption (Ghose and Han 2011b), Internet usage and mobile Internet characteristics (Ghose and Han 2011a), users' browsing behavior on mobile phones and personal computers (Ghose, Goldfarb, and Han 2012), and voice and short message price elasticity (Kim et al. 2010). Although these studies represent attempts to illuminate the nature of the mobile app market, none has extracted the effect of social dynamics on consumers' choices in the context of the mobile app store, at both the macro and micro levels.

*Global Macro and Micro Diffusion and Social Learning*

Two main streams of literature in product diffusion are relevant to this study: micro diffusion models and the global diffusion and social learning models. The earliest micro diffusion model considers consumers' Bayesian learning from the signals that follow a Poisson process (Chaterjee and Eliashberg 1990). Later studies emphasize the need for micro diffusion modeling (Young 2009) and critically review the aggregation and homogeneity of diffusion models (Peres, Muller, and Mahajan 2010). To remedy the issues, some studies propose micro network topology approaches (Dover, Goldenberg, and Shapira 2012; Iyengar and Van den Bulte 2011). Other studies suggest structural modeling of consumers' dynamic, forward-looking adoption choices (Song and Chintagunta 2003) and systematic conditioning to heterogeneous consumers' adoption choices (Trusov, Rand, and Joshi 2013). Peres, Muller, and Mahajan (2010) present a review of this literature stream. Parallel to the micro diffusion literature, a stream of studies has provided solutions for heterogeneous social learning processes (Van den Bulte and Joshi 2007), the mixing (interactions) of adoption processes (Putsis et al. 1997), simultaneous diffusion (Dekimpe, Parker, and Sarvary 1998), supply-side relationship (e.g., production economies) and omitted variable (e.g., income) correlations (Putsis and Srinivasan 2000), and the effect of macroenvironmental variables (Talukdar, Sudhir, and Ainslie 2002).

The current study builds on this literature by proposing a prescriptive machine-learning pipeline that combines the advantages of both macro and micro modeling approaches. The approach proposed recognizes that the app store platform's data may be a noisy measure of the variables of interest. We deal with the sparsity of choices through a combination of aggregation, filtering, hierarchy, and SUR processes. We suggest data-cleaning and modeling approaches that may be suitable for the big data variety, velocity, veracity, and volume of the app store platform. To estimate the model, we also suggest a genetic optimization metaheuristic approach, which enables the stochastic surface optimization.

*MODEL DEVELOPMENT*

We begin the modeling section with the choice of consumers $(i = 1, ..., I)$ at the app store. However, to recognize the long-tail distribution of mobile app choices (which creates sparsity), we aggregate the choice data at the app category level. The customer makes a choice of mobile app category j $(j = 0,1,...,J)$ in a given week $(t = 0,1,...,T)$, where $j = 0$ denotes the outside good option. The model of consumer app choice is different from that in prior studies (Carare 2012), in that instead of modeling aggregate purchases, it models individual-specific choices through a rich set of mobile app category characteristics. The model is similar to the nested logit model structure in studies such as Ratchford (1982) and Kök and Xu (2011), but to recognize sparsity of end choices, we aggregate the choices within the nest as the choice of the nests. This model may be useful for mobile app store owners concerned about the diffusion of mobile apps within the categories rather than the diffusion of each instance of mobile app in seclusion.

We specify the utility of consumers' choice of app categories in the app store in the following form:

$$u_{ijt} = \alpha_{ij} + \alpha_{i11}s_{it} + \alpha_{i12}\hat{c}_{jt}^{imm} + \alpha_{i13}F_{1jt} + \alpha_{i14}F_{2jt} + \alpha_{i15}F_{3jt} + \varepsilon_{it}^{j},$$

(1)

where $\alpha_{ij}$ denotes the random coefficient of consumer i's preference for mobile app category j; $F_{1jt}$, $F_{2jt}$, and $F_{3jt}$ denote factors that control for variation in observable mobile app characteristics/quality, price, and advertising, respectively; $\hat{c}_{jt}^{imm}$ denotes a time-varying social influence measure; and $s_{it}$ denotes the history of consumer i's app downloads until time t, which controls for state dependence and app-choice interdependence. In particular, if consumer i downloads an app at t − 1, then $s_{it} = s_{it-1} + 1$; otherwise, if the consumer selects an outside option, then $s_{it}$ remains unchanged (i.e., $s_{it} = s_{it-1}$). This specification induces a first-order Markov process on the choices. Controlling for state dependence and social influence helps consider the potential correlation between customers' choices across the categories and across the individuals. Table 2 presents the definition of the variables and the parameters.

-- Table 2. here --

Assuming that the random utility term $\varepsilon_{it}^j$ has type I extreme value distribution, consumer $i$'s probability of selecting the app category $j$ at time $t$ is given by a multinomial logit model, based on the deterministic portion $v_{ijt}$ of random utility $u_{ijt}$ as follows:

$$p_{ijt} = \frac{\exp(v_{ijt})}{1 + \sum_{j=1}^{J} \exp(v_{ijt})},$$ (2)

where the mean utility of the outside good is set to zero (i.e., $v_{it}^0 = 0$). A vector of mobile app category characteristics includes the average file size of mobile apps (a proxy for the app quality), frequency of featuring, average and variance of mobile app prices, the number of paid or free apps and their ratio, and the average tenure (time since creation) of all the mobile apps in the category. These variables can act as measures of (proxy for) competition. We assume that each of these pieces of the data contains some information that may be important for the consumer, but these pieces are highly correlated. Therefore, to gain better insight, we reduce the variation in these variables to three factors that preserve 85% of the variation. Formally, we use the following factor model process:

$$x_{jt} = bF_{jt} + e_{jt}^{'}, e_{jt}^{k'} \sim N(0, E),$$ (3)

To model consumer social learning, we use the filtered latent time-varying density of imitators $\bar{c}_{jt}^{imm}$. This approach is similar to the classic practice of modeling consumers' responses to featured and display products, in which the modeler includes an aggregate measure into the choice model to measure responses. Furthermore, the theoretical interpretation of this modeling approach is that as the number of imitators in the population increases, the possibility that an individual observes another individual who has already adopted the mobile app also increases. As a result, the consumer may become more or less likely to adopt a mobile app within mobile app category $j$. This theory is similar to the micro modeling diffusion proposed by Chatterjee and Eliashberg (1990), except that the model does not assume that the consumer receives information with a Poisson process, so the process can be a nonhomogeneous Poisson process (interarrival time is no longer memoryless). In other word, we endogenize consumers' information-receiving process in the choice model. This approach serves as an alternative to the micro modeling approach Yang and Allenby (2003) use to incorporate interdependence of awareness and preferences of consumers, but this model is useful when micro spatial structure information is not available. Our proposed approach may be relevant to the context of pervasive goods because these goods are more visible in daily interactions.

Two approaches can capture the density of imitators in the model. The first approach is to model density as a latent-state variable and recover it from the choice model. Though fancy, this approach may not be the best approach over big data, because it is computationally intractable. The second approach is to use the aggregate diffusion data to filter out the number of imitators. This approach combines macro aggregate diffusion modeling with micro choice modeling methods, to endogenize the number of imitators, an approach that may be more suitable for big data. In this approach, we can use aggregate diffusion data to filter out the number of imitators with a two-degree polynomial linear model. Then, we can use the filtered data in the choice model to run a nonlinear model on the data set of consumers' individual choices.

We use the whole data set to filter out the density of influentials and imitators for the mobile app category $j$ within the population at the given time $t$. We cast the social learning diffusion differential equations (Van den Bulte and Joshi 2007) into a discrete state-space model. This model is similar to a double-barrel Bass diffusion model and allows for heterogeneity in the adopters, by segmenting the observed cumulative number of adopters into the latent number of imitators and influentials. In contrast with classic log-likelihood and nonlinear least squares methods, our filtering approach increases estimation robustness to process and measurement noises (Xie et al. 1997).

$$y_{jt} = \theta_j c_{jt}^{Inf} + (1 - \theta_j) c_{jt}^{imm} + v_{jt}, v_{jt} \sim N(0, V_j)$$

$$\dot{c}_{jt}^{Inf} = (p_j^{\inf} + q_j^{\inf}(\frac{c_{jt-1}^{\inf}}{M_j^{\inf}}))(M_j^{\inf} - c_{jt-1}^{\inf}) + e_{jt}^{\inf}, e_{jt} \sim MVN(0, W)$$

$$\dot{c}_{jt}^{imm} = (p_j^{imm} + q_j^{imm}(w_j(\frac{c_{jt-1}^{\inf}}{M_j^{\inf}}) + (1 - w_j)(\frac{c_{jt-1}^{imm}}{M_j^{imm}})))(M_j^{imm} - c_{jt-1}^{imm}) + e_{jt}^{imm} , \qquad (4)$$

where $y_{jt}$ denotes the observed cumulative number of adopters of mobile apps in the mobile app category j at time (day) t; $c_{jt}^{\inf}$ denotes the latent cumulative number of adopters in the influential segment for app category j at time (day) t; $c_{jt}^{imm}$ denotes the latent cumulative number of adopters in the imitator segment for app category j at time(day) t; $\theta_j$ denotes the size of the segment of influential adopters, bound between zero and one; $p_j^{\inf}$ denotes the independent (random) rate of adoption of influential adopters; $q_j^{\inf}$ denotes the dependent (influenced by other influential adopters) rate of adoption of influential adopters; $p_j^{imm}$ denotes the independent (random) rate of adoption of imitator adopters; $q_j^{\inf}$ denotes the dependent (influenced by other adopters) rate of adoption of imitator adopters; $w_j$ denotes the degree of influence of influential adopters on the adoption of imitators; $v_{jt}$ denotes the noise of observation equation; and $(e_{jt}^{\inf}, e_{jt}^{imm})$ denotes the vector of noises of state equations.

In summary, Equation 1 denotes the observation equation and Equations 2 and 3 the state equations of the state-space model. Equation 1 uses a discrete latent model to integrate over the cumulative number of influential and imitator adopters. Equation 2 captures the adoption process of the influential adopters segment, while Equation 3 captures the adoption process of the imitator adopters segment. The imitators are different behaviorally from the influentials, in that they learn not only from themselves but also from consumers in the influential segment.

This model of social influence measure is more suitable for the context of mobile apps, as it better captures the social learning process (Van den Bulte and Joshi 2007) than the information cascade process (Bass 1969). Furthermore, it allows for heterogeneity in the adoption process, by segmenting the adopters into influential and imitator segments. Van den Bulte and Joshi (2007) find a closed-form solution for this model; however, we suggest that they may have measured the data with noise. As a result, to control for this potential measurement error, we use a state-space model structure with observation and state noises.

We recognize that there is shared information in the diffusion of various mobile app categories in the app store. Therefore, we model these differential equations of social learning across mobile app categories jointly and simultaneously. This joint modeling captures shared information at two levels: covariance and prior.

To account for simultaneity, on the covariance level, we model the state variance of the latent measure of cumulative influential and imitator adopters and the variance of the state equation of cumulative influential and imitator adopters through a SUR model. The SUR model presented formally in Equation 4 models the joint distribution of the state equations in a multivariate normal model structure, rather than modeling the state equation error terms individually.

To jointly model the diffusions, we use a hierarchical model (prior) with a conditionally normal distribution constraint on the fixed app-category-specific diffusion parameters, which is $\Phi_j = (p_j^{\text{inf}}, p_j^{imm}, q_j^{\text{inf}}, q_j^{imm}, M_j^{\text{inf}}, M_j^{imm}, \theta_j, w_j)$. This Bayesian process shrinks the fixed app-category-specific parameters toward the popularity of each mobile app, because we expect that more popular mobile apps have higher rates of imitator adoptions and market size. Formally, we define the following structure:

$$\Phi_j = \Delta_o Pop_j + o_j, o_j \sim N(0, \sigma_o^2),$$  (5)

where $\Phi_j$ denotes a vector of nonstate (fixed) parameters of the diffusion; $Pop_j$ denotes the popularity of mobile app category j; $\Delta_o$ denotes the hyperparameter of app-category-specific parameter shrinkage; and $o_j$ denotes the noise of the hierarchical model, or the unobserved heterogeneity of the mobile app categories.

We account for heterogeneity in the individual choice parameters by modeling the choice parameters' random effects. To consider the possibility of misspecification that may result from a rigid normal prior, we adopt the flexible semiparametric approach Dubé et al. (2010) propose. This approach assumes a mixture of multivariate normal distributions over the parameters' prior, to allow for a thick tail–skewed multimodal distribution. We denote the vector of fixed consumer-level parameters by $A_i = (\alpha_{i1}, \alpha_{i2}, ..., \alpha_{i15})$. We accommodate consumer heterogeneity by assuming that $A_i$ is drawn from a distribution common across consumers, in two stages. We employ a mixture of normal as the first stage prior, to specify an informative prior that also does not overfit. The first stage consists of a mixture of $K$ multivariate normal distributions, and the second stage consists of a prior on the parameters of the mixture of normal density. Formally:

$$p(A_i - \Delta z_i \mid \pi, \{\mu_k, \Sigma_k\}) = \sum_{k=1}^{K} \pi_k \phi(A_i - \Delta z_i \mid \mu_k, \Sigma_k)$$
$$\pi, \{\mu_k, \Sigma_k\} \mid b$$  (6)

where $b$ denotes the hyperparameter for the priors on the mixing probabilities and the parameters governing each mixture component; $K$ denotes the number of mixture components; $\{\mu_k, \Sigma_k\}$ denotes a mean and covariance matrix of the distribution of the individual-specific parameter vector $A_i$ for mixture component k; $\pi_k$ denotes the size of the $k$th component of the mixture model; $\phi$ denotes the normal density function distribution; $z_i$ denotes the information set on customer i, which here includes only tenure (the number of days from customer $i$'s registration on the app store); and $\Delta$ denotes the parameter of correlation between the choice response parameter and the information set on customer $i$.

Obtaining a truly nonparametric estimate using the mixture of normal model requires that the number of mixture components $K$ increases with the sample size. We adopt the nonparametric Bayesian approach Rossi (2014) proposes, which is equivalent to the approach mentioned previously when $K$ tends to infinity. In this structure, the parameters of the mixture normal model have a Dirichlet process prior. The Dirichlet process is the generalization of a Dirichlet distribution for an infinite atomic number of partitions. This process represents the distribution of a random measure (i.e., probability). The Dirichlet process has two parameters: the base distribution, which is the parameters of the multivariate Normal-inverse Wishart (N-IW) conjugate prior distribution for the distribution of the partitions from which the choice parameters are drawn, and the concentration parameter. Formally, the prior for the individual-specific choice parameters has the following structure:

$$\theta_{k1} = (\mu_{k1}, \Sigma_{k1}) \sim DP(G0(\lambda), \alpha^d)$$
$$G0(\lambda): \mu_{k1} \mid \Sigma_{k1} \sim N(0, \Sigma_{k1} a^{-1}), \Sigma_{k1} \sim IW(v, v \times v \times I)$$
$$\lambda(a, v, \upsilon): a \sim Unif(\bar{a}, \bar{a}), v \sim d - 1 + \exp(z), z \sim Unif(d - 1 + \vec{v}, \vec{v}), \upsilon \sim Unif(\vec{\upsilon}, \vec{\upsilon})$$
$$\alpha^d \sim (1 - (\alpha - \vec{\alpha})/(\vec{\alpha} - \vec{\alpha}))^{power}$$

$$(7)$$

where $G0(\lambda)$ denotes the base distribution or measure (i.e., the distribution of hyperparameters of the prior distribution of the partitions); $\lambda$ denotes the random measure, which represents the probability distribution of $(a, v, \upsilon)$; $(a, v, \upsilon)$ denotes the hyperparameters of the prior distribution of the partitions to which the choice parameters belong, which represent the behavior parameters of the latent segments; $d$ denotes the number of choice parameters per customer (in this case, d is equal to 15); and $\alpha^d$ denotes the concentration (also referred to as precision, tightness, or innovation) parameter. The idea is that the Dirichlet process is centered over the base measure $G0(\lambda)$ with N-IW with precision parameter $\alpha^d$ (a larger value denotes tight distribution). The term $(\vec{a}, \vec{a}, \vec{\nu}, \vec{\nu}, \vec{\upsilon}, \vec{\upsilon})$ denotes the hyperparameter vector for the second-level prior on the hyperparameters of prior over the partition distribution of the choice parameters.

The Dirichlet process mixture captures the distribution over the probability measure defined on some sigma-algebra (collection of subsets) of space $\aleph$, such that the distribution for any finite partition of $\aleph$ is a Dirichlet distribution (Rossi 2014). In this case, the probability measure over the partitions for the mean and variance of random coefficient response parameters of individual choice parameters sigma-algebra has the N-IW conjugate probability. For any subset of customers $C$ of $\aleph$:

$$E[G(C_\lambda)] = G_0(A_\lambda)$$
$$Var(G(C_\lambda)) = \frac{G_0(A_\lambda)(1 - G_0(A_\lambda))}{\alpha^d + 1},$$

$\qquad(8)$

According to the De Finetti theorem, integrating (marginalizing) out the random measure $G$ results in the joint distribution for the collection of the individual-specific mean and covariance of random coefficient choice parameters as follows:

$$p(\mu, \Sigma) = \int p(\mu, \Sigma \mid G) p(G) dG.$$

$\qquad(9)$

This joint distribution can be represented as a sequence of conditional distributions that has exchangeability property

$$p((\mu_1, \Sigma_1), ..., (\mu_I, \Sigma_I)) = p((\mu_1, \Sigma_1)) p((\mu_2, \Sigma_2) \mid (\mu_1, \Sigma_1)) ... p((\mu_I, \Sigma_I) \mid (\mu_1, \Sigma_1), ..., (\mu_{n-1}, \Sigma_{n-1})),$$

(10)

The Dirichlet process is similar in nature to the Chinese restaurant process and Polya Urn. In the Chinese restaurant process, there is a restaurant with an infinite number of tables (analogous to partitions of mean and variance of the individual choice random coefficients). A customer entering the restaurant selects a table randomly, but he or she selects the table with probability proportional to the number of customers who have sat at that table so far (in which case, the customer behaves similar to the other customers who have sat at the selected table). If the customer selects a new table, he or she will behave according to a parameter that he or she randomly selects from other restaurant customer behavior parameters (so not necessarily identical to the parameters of the other tables). The Polya Urn process also has the same structure. In this process, the experimenter begins by drawing balls with different colors from the urn. Any time the experimenter draws a ball with a given color, he or she adds an additional ball with the same color to the urn and also returns the drawn ball. The distribution of the number of customers sitting at each table in the Chinese restaurant process and the number of balls in each color in Polya Urn follow the Dirichlet process. An alternative way is Dubé et al.'s (2010) approach to fit models with successively large numbers of components and to gauge the adequacy of the number of components by examining the fitted density associated with the selected number of components. However, the process of model selection is tedious in this case.

-- Insert Figure 1 here --

*DATA*

The data set was collected by an African telecom operator on individual choices of downloading mobile apps from the app store platform of its global partner. The app stores are a type of two-sided platform, as they match consumers and developers/publishers, without considering the ownership of the mobile apps. The app store was launched approximately 330 days before commencement of the study in 2013 and 2014. We used the aggregate download data for a period of around 190 to 259 days as the macro sample and the data on download choices of a sample of 1258 consumers for a period of 124 days as the micro sample. The macro sample therefore includes between 1900 and 2590 observations; although this might not be considered big, the small sample includes approximately 160,000 observations, which are too big for nonlinear models.

A big data set such as ours creates a trade-off in estimation. On the one hand, big data can give insight into a short planning horizon, given our use of a linear model. On the other hand, computationally intensive methods can give insights into prescriptive power, given that the data are not big. We wanted to employ a method that gave us the advantage of both big data and a computationally intensive method. As a result, we used a second-degree polynomial macro model of social learning diffusion over the macro sample and the nonlinear computationally intensive micro choice model over the micro sample.

To deal with the sparsity of the data, which is driven by the long-tail distribution of the mobile apps' adoptions, and to reduce the daily noise in the data, we aggregated the data of the micro sample at a weekly level before entering them into the choice model. We also aggregated the macro app adoption and micro app download choice data at the app category level to limit the study to the topic of interest for the app store platform, as well as to handle the data volume. In addition, we used a flexible Bayesian prior to shrink the individual-specific choice parameters. We investigated two sources of consumer preference interdependence: local and global. For the local interdependence, we filtered out the macro sample data to individual adopters who live in the city of interest. To do this, we used a data set of mapping IP addresses to cities we collected by crawling the World Wide Web. For the global interdependence, we used the aggregate information about the mobile app adoptions in all 30 cities.

-- Table 3. here --

In a nutshell, the data consist of approximately 20,000 consumers, with around 3000 consumers in the local African city studied. This local city had approximately 4000 app downloads for the duration of the study. These figures classify the current data as big, in terms of variety, velocity, veracity, and volume. Table 3 illustrates the list of categories selected and their corresponding total downloads in the local city. Each of the 1258 customers adopted only one of the mobile apps during the course of study, so on all other days they selected an outside option. This observation suggests that a mixed logit choice model might be suitable, but only if the intertemporal dependencies between the choices are controlled.

-- Insert Figure 2 here --

The data set also included longitude and latitude of each IP address. However, as mobile phones are usually attached to customers who might move within the city, aggregating locations at the city level might be relevant. Moreover, this assumption is innocuous because of the social nature of mobile phones and mobile apps (i.e., usage of mobile apps and mobile phones in a social atmosphere). Mobile phones have become ubiquitous, and mobile phone use at social events makes mobile apps visible, thus creating social learning opportunities. Figure 2 plots the diffusion curves of the cumulative adoptions of a sample of six mobile app categories.

For each mobile app category, we had the average file size, the total number of apps featured, the average and the variance of app prices, and the number of paid and free options. Table 4 presents the basic statistics of these variables. To explain the heterogeneity in individual responses, we used the data on the tenure of each customer, or the number of days since each customer had subscribed to the app store. As different types of consumers (i.e., influentials and imitators) with different psychological traits adopt the technology at different points in time (Kirton 1976), we used the tenure of consumers as a proxy for the psychological traits that can explain the heterogeneity in their choice responses.

-- Table 4 here --

To explain heterogeneity in app store categories, we used the popularity of the mobile app categories in the Apple app store. We expected that the popularity of the mobile app categories on the Apple app store would explain the diffusion of the mobile app categories on the other app store platforms as well. Figure 3 presents the popularity statistics; this figure shows the long-tail distribution of the popularity of the mobile apps.

-- Figure 3 here --

Figure 4, from Distimo, a mobile app market research company, suggests that some mobile app categories are more likely to be paid and others are more apt to be free. The high share of free mobile apps is an important observation. The same feature exists in the data sets used in this study. This feature implies that the key cost factor consumers incur might be the cost of learning about the application, supposedly from others (e.g., friends, over the Internet). This observation suggests that social influence is an important factor for adoption decision, but a formal model is required to confirm this conjecture.

-- Figure 4 here --

A mental accounting perspective would suggest that consumers perceive paid mobile apps as investments, for which they are willing to pay money, and free mobile apps as entertainment (Chang 2014), for which they might be less inclined to pay. Guided by the same intuition, we classified mobile app categories in the sample into two categories: utilitarian and hedonic. The utilitarian category includes device tools, health/diet/fitness, Internet/Wireless Application Protocol (WAP), and reference/dictionary mobile apps. These types of apps might be prominent for their utility rather than their entertainment. The hedonic category includes eBooks, games, humor/jokes, logic/puzzle/trivia, and social network mobile apps. These types of apps might be more relevant for their entertainment features. This categorization allows for investigation of whether customers value the utilitarian mobile app categories more than the hedonic mobile app categories, based on the customer choice parameters.

Finally, the mobile apps are more similar to durable goods than to nondurables. Therefore, consumers' choice of downloading a mobile app may be sparse in nature. Sparsity here means that consumers' other choices are to not download or to choose outside options. A suitable modeling approach that can handle this sparsity is hierarchical Bayes, which borrows information from other sample items to overcome the sparsity issue.

*IDENTIFICATION AND ESTIMATION*

To identify the choice model, we used a random coefficient logit specification, which has a fixed diagonal scale. To set the location of the utility, we normalized the utility of the outside option to zero. To minimize concerns about endogeneity (omitted variable), we control for potential correlations between choices by explicitly modeling the intertemporal choice interdependence in the choice history state variable. We also control for potential confounding effects of price, advertising, and product characteristics by including the latent factors of variation in these variables in the choice model. To control for potential measurement error in the social influence measure, we use UKF, and we control for potential simultaneity in the social measures through a SUR model structure. In addition, by using a random coefficient structure, the modeling approach minimizes any concern with independence from irrelevant alternatives, as it allows for heterogeneity in the individual-specific choice behavior parameters.

We identified individual-level choice parameters using the micro sample panel of consumers, a sample that consists of 20-week micro choices of 1258 customers. Bayesian shrinkage with a flexible Dirichlet process prior helps identify the large set of individual-specific parameters, without overfitting. The mixture normal distribution is subject to a label-switching problem (i.e., the permutation of segment assignment returns the same likelihood). However, we avoided this problem by limiting inference to the joint distribution rather than individual segment assignment. To estimate the micro choice model on the micro sample, we used multinomial logit with Dirichlet process prior on the individual-specific hyperparameter (Bayesian semiparametric) estimation code from Bayesm package in R. This method uses the Metropolis–Hasting random walk (MH-RW) method to estimate conditional choice probabilities on cross-sectional units (i.e., customers). The limitation of MH-RW is that random-walk increments are tuned to conform as closely as possible to the curvature in the individual-specific conditional posterior, formally defined by:

$$p(A_i \mid y_i, \mu, \Sigma, z_i, \Delta) \propto p(y_i \mid A_i) p(A_i \mid \mu, \Sigma, z_i, \Delta)$$ (11)

Without prior information on highly probable values of a first-stage prior (i.e., $p(A_i \mid .)$), tuning the Metropolis chains given limited information of cross-sections (i.e., each customer) by trial is difficult (this problem is exacerbated when customers do not have some of the choice items selected in their history). Therefore, to avoid singular hessian, we implemented the fractional likelihood approach of Rossi, Allenby, and McCulloch (2005) in our approach. Formally, rather than using individual-specific likelihood, the MH-RW approach forms a fractional combination of the unit-level likelihood and the pooled likelihood as follows:

$$l_i * (A_i) = l_i (A_i)^{(1-w)} \left( \prod_{i=1}^{I} l_i (A_i \mid y_i) \right)^{w\beta}, \beta = \frac{n_i}{N}, N = \sum_{i=1}^{I} n_i,$$ (12)

where $w$ denotes the small tuning parameter to control for the effect of pooled likelihood $\prod_{i=1}^{I} l_i (A_i \mid y_i)$, $\beta$ denotes a parameter chosen to properly scale the pooled likelihood to the same order as the unit likelihood, and $n_i$ denotes the number of observations for customer $i$. Using this approach, the MH-RW generates samples conditional on the partition membership indicator for consumer $i$ from proposal density $N(0, s^2\Omega)$, such that

$$\Omega = (H_i + V_A^{-1})^{-1}, H_i = -\frac{\partial^2 \log l_i *}{\partial A \partial A'}\big|_{A=\hat{A}_i},$$ (13)

where $\hat{A}_i$ denotes the maximum of the modified likelihood $l_i * (A_i)$ and $V_A$ denotes a normal covariance matrix assigned to the partition (i.e., segment) to which customer $i$ belongs.

This approach considers that $A_i$ is sufficient to model the random coefficient distribution. To estimate the infinite mixture of the normal prior for choice parameters, a standard data augmentation with the indicator of the normal component is required. Conditional on this indicator, we can identify a normal prior for the parameters of each customer $i$. The distribution for this indicator is multinomial, which is conjugate to a Dirichlet distribution. Formally:

$$\pi \sim Dirichlet(\alpha^d)$$
$$z^i \mid \pi \sim Mult-Nom(\pi).$$ (14)

As a result, we can define posterior by

$$z^i \sim Mult-Nom(\frac{\alpha^1}{\sum_j \alpha^j}, ..., \frac{\alpha^\kappa}{\sum_j \alpha^j})$$
$$\pi \mid z^i \sim Dirichlet(\alpha^1 + \delta_1(z^i), ..., \alpha^\kappa + \delta_\kappa(z^i)),$$ (15)

where $\delta_j(z^i)$ denotes whether or not $z^i = j$. This result is relevant for the Dirichlet process, as any finite subset of customers' choice-behavior parameters' partitions has Dirichlet distribution, and a finite sample can only represent a finite number of partitions. The exchangeability property of partitions allows for the use of an estimation approach to sequentially draw customer parameters, given the indicator value as follows:

$$(\mu_i, \Sigma_i) \mid (\mu_1, \Sigma_1),...,(\mu_{i-1}, \Sigma_{i-1}) \sim \frac{\alpha^d G_0 + \sum_{j=1}^{i-1} \delta_{(\mu_j, \Sigma_j)}}{\alpha + i - 1}. \tag{16}$$

The next portion of this approach's specification is the definition of the size of the finite clusters over the finite sample that is controlled by $\pi$. Rossi (2014) suggests augmenting Sethuraman's stick-breaking notion for draws of $\pi$. In this notion, a unit-level stick is iteratively broken from the tail with proportion to the draws with beta distribution with parameters one and $\alpha^d$, and the length of the broken portion defines the $k$th element of the probability measure vector $\pi$ (a form of multiplicative process). Formally:

$$\pi_k = \beta_k \prod_{i=1}^{k-1}(1 - \beta_i), \beta_k \sim Beta(1, \alpha^d). \tag{17}$$

In this notion, $\alpha^d$ determines the probability distribution of the number of unique values for the Dirichlet process mixture model, by

$$\Pr(I* = k) = \left\| S_i^{(k)} \right\| (\alpha^d)^k \frac{\Gamma(\alpha^d)}{\Gamma(i + \alpha^d)}, S_i^{(k)} = \frac{\Gamma(i)}{\Gamma(k)}(\gamma + \ln(i))^{k-1}, \tag{18}$$

where $I*$ denotes the number of unique values of $(\mu, \Sigma)$ in a sequence of $i$ draws from the Dirichlet process prior, $S_i^{(k)}$ denotes the Sterling number of the first kind, and $\gamma$ denotes Euler's constant. Furthermore, to facilitate assessment, this approach suggests the following distribution for $\alpha^d$, rather than Gamma distribution:

$$\alpha^d \propto (1 - \frac{\alpha^d - \bar{\alpha}}{\bar{\bar{\alpha}} - \bar{\alpha}})^\phi ,$$

(19)

where $\bar{\alpha}$ and $\bar{\bar{\alpha}}$ can be assessed by inspecting the mode of $I^* | \alpha^d$ and $\phi$ denotes the tunable power parameter to spread prior mass appropriately. An alternative to the Gibbs sampler employed by this approach is a collapsed Gibbs sampler, which integrates out the indicator variable for partition (segment) membership of each customer, but Rossi (2014) argues that such an approach does not improve the estimation procedure. Appendix C presents the series of conditional distributions that this approach employs in its Gibbs sampling to recover individual-specific choice parameters.

We identified the latent cumulative number of influentials and imitators of mobile app categories with the observed cumulative number of adopters in the complete data set. To avoid overfitting, we also used a normal prior on the fixed social learning macro diffusion model to regularize the likelihood of the model. Although the local and global aggregate data sets have only 2000 observations, Bayesian shrinkage of parameters allows identification of the parameters. To estimate the latent cumulative number of influentials and imitators, we used the MAP method, a popular method in machine learning, as an alternative method to MCMC sampling methods. This approach uses an optimization method to maximize the a posteriori of the model parameters. We used genetic algorithm for the optimization, as the number of parameters estimated for the social learning diffusion model is around 300: 210 covariance elements of state covariance matrix, 10 elements of observation covariance matrix, and 80 elements of fixed parameters of the diffusion differential equations. The gradient descend optimization method has complexity of $O(P)$ per iteration but requires a tuning learning parameter, and the Quasi-Newton optimization method has the complexity of $O(P^2)$ per iteration, where P is the number of parameters to estimate. This complexity translates to a long run time over big data, in which the number of parameters increases with the variety and volume of the data. As a result, we adopted the genetic algorithm approach, which Venkatesan, Krishnan, and Kumar (2004) find comparable to the classic gradient descend or the Quasi-Newton approach. In addition, genetic algorithm is known as a global optimization method, in contrast with local optimization of the Quasi-Newton method. Given that a latent-state-space model (e.g., mixture models) has multiple local maxima, a genetic algorithm is more likely to find the global maxima than a Quasi-Newton method.

To estimate the macro social learning diffusion model, we used the UKF nested within an MCEM method. The UKF is an approach proposed in robotics literature (Julier and Uhlmann 1997; Wan and Merwe 2001), which achieves third-order accuracy in estimating the latent state in a state-space model, as opposed to the extended Kalman filter, which only achieves first-order accuracy, with the same order of computational complexity (i.e., $O(T)$). The basic idea behind UKF is that rather than using the closed-form, first-order, tailor expansion term for the measurement updating of the latent state, by computing a Jacobean vector, it uses an unscented transformation to transform sigma vector of points around the mean and the mean of the latent-state prior of a nonlinear state equation to estimate the transformed normal distribution posterior parameters. We explain the UKF algorithm in Appendix B.

The MCEM approach begins with an initial vector of parameters. Then, it uses MCMC, UKF in this case, to recover the latent-state distribution and a set of samples. Given the latent-state samples, it computes the expected log-likelihood and searches for the parameters that maximize this expected log-likelihood (De Valpine 2012). MCEM is appealing for its speed, compared with the full MCMC sampling method. However, the MCMC approach is notorious for slow convergence, and both approaches may suffer from finding only the local maxima. The exercise of a global optimization genetic algorithm stochastic search may be a remedy to this stochastic surface search problem. In the optimization, we used transformation to ensure that the market sizes of the social learning diffusion model were positive and that parameters of the effect of learning from influentials and imitators in the imitator state equation ($w$) and the segment size of influentials and imitators ($\theta$) were between zero and one. We used the just-in-time compiler in R to speed up the estimation process.

*RESULTS*

Table 5 presents the log-likelihood of the proposed models. Models 1 and 2 represent social learning aggregate diffusion models over local adoption (only adopters in the local city) and global adoption (total number of adopters across 30 cities). The local social learning model dominates the global social learning model by the likelihood. This result suggests that the mobile app adoption process is more locally rather than globally coordinated. Models 3 and 4 use the filtered number of imitators as a measure for social influence, and Models 5 and 6 use the observed number of adopters as a measure for social influence. Domination of Models 5 and 6 over Models 3 and 4 by log-likelihood suggest that not only the number of imitators but other social factors as well are the driving factors for mobile app adoptions. These other factors might include the social force of differentiators (the result of the micro analysis confirms the existence of such a potential). The dominance of Models 5 and 6 over Model 7 (the model with no social learning) suggests that social learning is an important force that drives individual mobile app adoption choices (we discuss the bias in the parameter estimates when social learning is ignored subsequently).

--Table 5 here --

Finally, the dominance of Model 5 over Model 6 reconfirms the result from the aggregate model that social learning at the local (in the city) rather than the global (e.g., over the Internet) level drives customers' adoption choices. This finding for mobile apps (as a form of pervasive good) contrasts with the findings on the adoption of traditional goods that emphasize the importance of learning over the Internet (Putsis et al. 1997).

Figures 5 and 6 present a step-ahead forecast versus the observed cumulative number of adopters at both the local and global level. This visualization, together with the mean absolute deviation and the mean square error presented in Table 6, suggests that the social learning macro diffusion model fits the app store platforms' macro app diffusion data reasonably well. We benchmarked the estimates with Van den Bulte and Joshi's (2007) study, and relative to the market size, the mean square error falls in a reasonably good range.

-- Table 6 and Figures 5 and 6 here --

Table 7 presents the result of the factor analysis to extract the latent factor of mobile app characteristics. We used Varimax rotation to interpret the factors and named the factors from both the supply and the demand side in Table 8.

-- Tables 7 and 8 here --

We limited the factor/principal components to three, as it captures .85% of the variation in the data. The number of paid and free mobile apps load highly onto the first factor, thus indicating the high demand for these mobile apps, which likely led app developers/publishers to develop many mobile apps. Furthermore, Ghose and Han (2014) use the average file size of a mobile app as a proxy for the quality of mobile apps, and this mobile app category feature also loads highly onto the first factor; thus, we label the first factor as popular mobile apps. The average and variance of the prices load highly onto the second factor; thus, we label the second factor as investment mobile apps. Finally, we refer to the third factor as freemiums, because the fraction of free mobile apps is much higher than that of paid mobile apps.

Tables 9 and 10 present the parameter estimates of the social learning diffusion models over the global (across the 30 cities with the app store) and local (in the city of interest) diffusion data. Across both diffusion data sets, the independent random adoption rate for consumers in the influential segment is not statistically significant across different mobile app categories, except for eBooks (which might be driven by assortment size). However, this rate is significantly higher than the dependent adoption rate for this segment, which suggests that the model is properly identifying the behavior of the segment of influentials. For the influential segment, the rate of independent adoption is close to the rate for the Everclear music CD that Van den Bulte and Joshi (2007) find. However, for this segment, the dependent rate of adoption is similar to the rate for foreign language CD adoption in that study. This result might be driven by the low search cost of the influential segment in the app store, which in turn leads influentials to learn less from others.

-- Tables 9 and 10 here --

For the imitator segment, in almost all the categories, the rate of independent adoption (M = .288) is greater than the rate of dependent adoption (M = .198). For this segment, the independent rate of adoption is significantly more than the rate for goods proposed in classic economy that Van den Bulte and Joshi (2007) report. This difference is likely driven by the low search cost of mobile apps for imitators. The dependent rate of adoptions is similar to the rate for Everclear music CD that Van den Bulte and Joshi (2007) report. For global adoption, across the mobile app categories, the weight of influentials in driving imitators' dependent choice of adoption is .009, which is similar to the same parameter for the Everclear music CD. However, this rate is .50 for local adoption, which is similar to the rate for a John Hiatt music CD (Van den Bulte and Joshi 2007). The size of the influential segment in the observed sample for global adopter data is .044 and .042 for local adopter data, which is similar to the rate for the Everclear music CD (Van den Bulte and Joshi 2007). These results might suggest that customers' adoption behavior for mobile apps is similar to music CD adoptions, except that the independent rate of adoptions for imitators are higher while the dependent rate of adoptions for influentials is lower, driven by lower search costs.

Table 11 summarizes the individual parameters distribution for the choice model that uses the local number of adopters (unfiltered density) as a proxy for social influence. The negative mean for the preference parameter for each mobile app category indicates higher preference of outside options for customers. In the city under study, customers prefer mobile apps in health/diet/fitness, games, Internet/WAP, and device tools relatively more than mobile apps in social network, eBooks, and humor/jokes categories. Regarding the Apple app store popularity statistics (see Figure 3, the surprising result is the high preference of customers for health/diet/fitness mobile apps. This information can help this app store better target its marketing communication messages by highlighting this mobile app category.

The mean for the distribution of the download history state parameter is negative and significant. This negative effect of history suggests that this app store is not doing well in retaining customers, perhaps because of its appearance and nonoptimal shopping shelf. However, the effect of social influence is positive and significant, which suggests that there is positive spillover (possibly because of an awareness effect) of adoption in the population.

-- Table 11 here --

Appendix D presents the same results table for the choice model with local imitators, models with global imitators/adopters, and a model with no social influence. The model with no social influence underestimates the preference for mobile apps in almost all the categories except for eBook, humor/jokes, reference/dictionary, and university. In addition, this model underestimates the effect of popularity, investment, and free characteristics of the mobile apps. In summary, the model that does not account for social influence returns biased estimates for the parameters.

-- Table 12 here --

Table 12 presents the correlation between customer tenure (number of days since registeration in the app store) and choice parameters of customers. Those who register early with the app store have higher preferences for mobile apps in Internet/WAP and university mobile app categories. This correlation is relevant because mobile influentials might be more interested in improving their performance-oriented apps. In addition, these customers are more sensitive to download history and social influence. This result aligns well with product life-cycle theories, which argue that if a product does not pass the acceptance of early adopters, it will fall into a chasm, leading to early failiure.

Figure 7 presents the distribution of choice parameters. This distribution has a heavy tail, which highlights the importance of allowing for flexible heterogeneity distribution for the choice parameters. Targeting is a relevant application of micro choice modeling for app stores. Table 13 presents the distribution of significance and sign of each of the choice parameters at the individual customer level. Knowing the distribution of negative and positive response helps the app store target 53 customers who do not prefer health/diet/fitness or social network mobile apps. This correct targeting might help improve the usability of the app store.

*COUNTERFACTUAL ANALYSIS*

The advantage of the individual-specific choice model for the app store platforms is that it allows estimating the implications of the social influence policy for total expected adoption by simulation. We ran three counterfactual scenarios using the estimated choice model by modifying the level of social influence. Furthermore, we used the estimated model to find the optimal dynamic level of social influence to maximize the diffusion over the app store platform. Formally, we solve the following optimization problem:

$$\max_{\{c_{jt}^{imm}\}} \sum_{t=1}^{T} \sum_{j=2}^{J} \sum_{i=1}^{I} \frac{\exp(u_{ijt})}{1 + \sum_{j=1}^{J} \exp(u_{ijt})}$$

$$c_{jt-1}^{imm} \leq c_{jt}^{imm}, \forall j, t \qquad\qquad . \qquad\qquad (20)$$

Table 14 presents the implications of each of these four policies. Surprisingly, shutting down the social influence improves the total expected adoptions of mobile apps on this app store platform. This further confirms that this platform does not have enough quality to retain its customers. However, an optimal social influence policy shows a 13.6% increase in total expected adoptions of the platform. This optimal policy decreases adoption of mobile apps in the reference/dictionary category but increases the expected adoption of mobile app categories in logic/puzzle/trivia, games, and device tools the most.

Table 15 presents the result of regressing the improvement under the optimal social influence policy on popularity rank of the mobile app category. The correlation between mobile app category popularity and the improvement under the optimal policy is positive and significant. This result suggests that more popular mobile app categories improve under the optimal policy. As such, this app store can improve its adoption by 13.6% if it can use social influence to increase the adoption of more popular mobile app categories. Finally, Figure 8 presents the social influence level for this optimal policy. This policy suggests an early increase in social influence by using a viral marketing campaign.

-- Table 15 and Figure 8 here –

*CONCLUSION*

In this paper, we developed an approach that combines a macro diffusion model with a micro choice model to allow app stores to target their customers, and we proposed the Dirichlet process to model customers' heterogeneity and UKF to estimate a social influence measure. Then, using a large data set from an African app store, we demonstrated that social influence is an important factor in determining customers' adoption choice. The results show that ignoring social influence in modeling customers' adoption can bias the choice parameters' estimates. Furthermore, the results indicate that social influence on mobile app adoption choices is effective locally (in the city under study) rather than globally (over the Internet). We benchmarked the mobile app adoption process against the same process for classic economy goods and found that the mobile app adoption process is similar to the process for music CDs.

Finally, we illustrated how the estimated model can be used to analyze a counterfactual scenario in which the app store platform optimizes its intervening social influence. The counterfactual analysis shows that if this app store runs viral marketing campaigns focusing on more popular mobile app categories, it can increase its total adoption by 13.6%. The study's modeling approach, proposed estimation method, and derived empirical insights should be of interest to both practitioners and scholars in academia.

**Table 1 Literature Position of this study**

| Stream of Study | Interdependence of consumer preference | App Store | Global micro/macro Simultaneous Diffusion |
|---|---|---|---|
| Current study | * | * | * |
| Yang and Allenby (2003); Stephen and Toubia (2010); Bell and Song (2007); Aral and Walker (2011); Nair et al. (2010); Bradlow et al. (2005); Hartmann (2010); Yang et al. (2006); Narayan et al. (2011); Kurt et al. (2011); Chung and Rao (2012); Choi et al. (2010). | * | - | - |
| Ghose and Han (2014); Carare (2012); Garg and Telang (2013); Liu et al. (2012);    Ghose et al. (2012); Ghose and Han (2011b); Ghose and Han (2011a); Kim et al. (2010). | - | * | - |
| Van den Bulte and Joshi (2007); Young (2009); Chatterjee and Eliashberg (1990); Putsis et al. (1997); Dekimpe et al. (2000); Neelamegham and Chintagunta (1999);  Talukdar et al. (2002); Gatignon and Robertson (1989); Takada and Jain (1991); Dover et al. (2012). | - | - | * |

**Table 2 Model Variable Definitions**

| Variable | Description |
|---|---|
| App Category Daily Download( $y_{jt}$ ) | Cumulative number of consumers who download an app in app category j up until a given day t |
| App Category Weekly Download Latent ( $c_{jt}^{\inf}$ ) | Latent cumulative number of consumers from influential segment, who download an app in app category j up until a given day t. Consumers from influential segment only learn from each other, and not from imitators. |
| App Category Weekly Download Latent ( $c_{jt}^{imm}$ ) | Latent cumulative numbers of consumers from imitator segment, who download an app in app category j up until a given day t. Consumers from imitator segment learn both from each other, and adopters in influential segment. |
| Segment size ( $\theta_j$ ) | A parameter between zero and one that define the size of the influential segment |
| Internal Market Force ( $p_j^{\inf}$ , $p_j^{imm}$ ) | The random Poisson rate of adoption of individuals in influential and imitator segment respectively. |
| External Market Force ( $q_j^{\inf}$ , $q_j^{imm}$ ) | The endogenized imitation rate of adoption of individuals in influential and imitator segment respectively. |
| Learning split ( $w_j$ ) | The degree to which the individuals in imitator segment learn from adopters in the influential segment |
| Market size  ( $M_j^{\inf}$ , $M_j^{imm}$ ) | The market size of individual in influential and imitator segments respectively. |
| Category hierarchy parameters $\{\mu_{k2}, \Sigma_{k2}\}$ | Parameter of locally weighted regression parameters of the hierarchical prior of app category diffusion parameters |
| Full covariance matrix of state equation( $W$ ) | Full covariance matrix of state equation of macro diffusion model, which may suggest complementarity or substitution. |
| Variance of observation equation ( $V_j$ ) | Variance of observation equation of macro diffusion model |
| Category data ( $x_{jt}$ ) | Category j characteristic data at day t, including Average file size, total number of adds featured in the category, average price, variance of price, paid app options, free app options, fraction of free to paid apps within the category, average tenure of each app category, total app options within the category |
| Category Factors( $F_{jt}$ ) | Reduced factors explaining the variation in category data |
| Factor loading of Category ( $b$ ) | Factor loading of data item j of category data vector |
| Consumer utility from app category ( $u_{ijt}$ ) | Consumer i's utility from selecting an app in app category j at week t |
| App category preference ( $\alpha_{ij}$ ) | App-category-specific preference of consumer i |
| Individual download history state ( $s_{it}^j$ ) | State of individual i's download history in a given category j until week t |
| $\alpha_{i11}...\alpha_{i15}$ | Utility parameters of consumer i |
| $p_{ijt}$ | Probability of selecting an app in category j at time t |
| $\pi_1, \{\mu_{k1}, \Sigma_{k1}\}$ | Parameter of hierarchical mixture of normal components of individual choice parameters |
| $v_{jt}, e_{jt}, e'_{jt}$ | Error terms of observation/state equation and factor model |

**Table 3 Categories Basic Statistics**

| Index | Category | Total Downloads within local city |
|---|---|---|
| 1 | Dating | 27 |
| 2 | eBook | 414 |
| 3 | Education & Learning | 24 |
| 4 | Health/Diet/Fitness | 42 |
| 5 | Internet & WAP | 52 |
| 6 | Movie/Trailer | 597 |
| 7 | POI/Guides | 22 |
| 8 | Reference/Dictionaries | 55 |
| 9 | TV/Shows | 135 |
| 10 | Video & TV | 105 |

**Table 4 Mobile app categories basic statistics**

| Category Data Summary | Mean | Variance | Min | Max |
|---|---|---|---|---|
| Number of available apps in the Category | 35 | 1250 | 12 | 141 |
| Average tenure of apps in the category (Days) | 316 | 6,386 | 169 | 498 |
| Number of available free apps in the category | 32 | 908 | 7 | 120 |
| Average days that an app is featured in the category | 0.12 | 0.05 | 0.00 | 0.71 |
| Average file size of apps in the category (MB) | 2.00 | 4.00 | 0.50 | 8.00 |
| Variance of prices of apps in the category | 0.51 | 1.09 | 0.00 | 3.75 |

**Table 5 MODEL COMPARISON**

| Model | Description | Number of obs. | Log Lik. |
|---|---|---|---|
| 1 | Local Adoption (aggregate sample) | 2,000 | -20,724.16 |
| 2 | Global Adoption (aggregate sample) | 2,000 | -21,649.32 |
| 3 | Choice Explained by Local Imitators Signals (micro sample) | 22,644 | -25,921.92 |
| 4 | Choice Explained by Global Imitators Signals (micro sample) | 22,644 | -38,310.49 |
| 5 | Choice Explained by Local Adopters Signals (micro sample)* | 22,644 | -12,252.85 |
| 6 | Choice Explained by Global Adopters Signals (micro sample) | 22,644 | -15,275.20 |
| 7 | Choice with No social influence measure (micro sample) | 22,644 | -15,977.04 |

\* dominant model

**Table 6 Performance of the Proposed Model for local and international category adoption**

| Description | Mean Absolute Deviation | Mean Square Error |
|---|---|---|
| Local Category Adoption | 0.64 | 1.48 |
| International Category Adoption | 0.03 | 0.12 |

**Table 7 Factor Loading Matrix (Varimax rotation)**

| Loadings/Components | C1 | C2 | C3 |
|---|---|---|---|
| Average File size (a proxy for app quality) | 0.77 | -0.07 | -0.09 |
| Dummy variable of Is Featured | 0.82 | 0.3 | 0.01 |
| Average Price | -0.06 | 0.94 | -0.28 |
| Variance of Price | -0.05 | 0.94 | -0.19 |
| Number of Paid app Options | 0.97 | -0.09 | -0.09 |
| Number of Free app Options | 0.96 | -0.15 | -0.03 |
| Fraction of Free apps to Paid Apps | -0.09 | -0.25 | 0.87 |
| Average Tenure (time from creation) | -0.08 | 0.67 | 0.48 |
| Total number of app Options | 0.96 | -0.14 | -0.03 |

**Table 8 Factor Names**

| Factor | Supply side Name | Demand Side Name |
|---|---|---|
| C1 | Red Ocean app categories | Popular Apps |
| C2 | Paid app categories | Investment Apps |
| C3 | Free app categories | Freemiums |

**Table 9 PARAMETER ESTIMATES: Global Adoption**

| | $p^{\text{inf}}$ | $q^{\text{inf}}$ | $p^{imm}$ | $q^{imm}$ | $M^{\text{inf}}$ | $M^{imm}$ | $w$ | $\theta$ |
|---|---|---|---|---|---|---|---|---|
| Mobile App Categories: | | | | | | | | |
| Device Tools | 0.024 | 0.000 | 0.278* | 0.192* | 50* | 580* | 0.010* | 0.039* |
| eBooks | 0.024 | 0.000 | 0.274* | 0.189* | 260* | 4540* | 0.007* | 0.044* |
| Games | 0.026 | 0.000 | 0.293* | 0.202* | 80* | 1150* | 0.009* | 0.046* |
| Health/Diet/Fitness | 0.025 | 0.000 | 0.288* | 0.199* | 100* | 1600* | 0.008* | 0.048* |
| Humor/Jokes | 0.026 | 0.000 | 0.297* | 0.205* | 100* | 1410* | 0.008* | 0.043* |
| Internet/WAP | 0.026 | 0.000 | 0.296* | 0.204* | 100* | 1580* | 0.010* | 0.039* |
| Logic/Puzzle/Trivia | 0.024 | 0.000 | 0.275* | 0.190* | 90* | 1440* | 0.009* | 0.039* |
| Reference/Dictionaries | 0.026 | 0.000 | 0.296* | 0.204* | 90* | 1440* | 0.007* | 0.048* |
| Social Networks | 0.026 | 0.000 | 0.297* | 0.205* | 40* | 390* | 0.008* | 0.044* |
| University | 0.025 | 0.000 | 0.281* | 0.193* | 130* | 2080* | 0.009* | 0.046* |

* $p \leq 0.05$

**Table 10 PARAMETER ESTIMATES: Local Adoption**

| Mobile App Categories: | $p^{\text{inf}}$ | $q^{\text{inf}}$ | $p^{imm}$ | $q^{imm}$ | $M^{\text{inf}}$ | $M^{imm}$ | $w$ | $\theta$ |
|---|---|---|---|---|---|---|---|---|
| Device Tools | 0.025 | 0.000 | 0.282* | 0.194* | 5* | 80* | 0.100* | 0.046* |
| eBooks | 0.024* | 0.000 | 0.278* | 0.191* | 103* | 1952* | 0.032* | 0.044* |
| Games | 0.024 | 0.000 | 0.275* | 0.189* | 3* | 56* | 0.246* | 0.038* |
| Health/Diet/Fitness | 0.025 | 0.000 | 0.282* | 0.194* | 7* | 120* | 0.782* | 0.038* |
| Humor/Jokes | 0.026 | 0.000 | 0.299* | 0.206* | 6* | 99* | 0.506* | 0.041* |
| Internet/WAP | 0.025 | 0.000 | 0.285* | 0.197* | 11* | 200* | 0.738* | 0.041* |
| Logic/Puzzle/Trivia | 0.025 | 0.000 | 0.282* | 0.194* | 6* | 113* | 0.344* | 0.043* |
| Reference/Dictionaries | 0.026 | 0.000 | 0.299* | 0.206* | 12* | 225* | 0.940* | 0.042* |
| Social Networks | 0.025 | 0.000 | 0.281* | 0.193* | 3* | 48* | 0.658* | 0.040* |
| University | 0.025 | 0.000 | 0.281* | 0.194* | 6* | 113* | 0.555* | 0.042* |

* $p \leq 0.05$

**Table 11 PARAMETER ESTIMATES: Individual Choice effect (Local Adopters)**

| | Estimate | Std. Dev. | 2.5$^{th}$ | 97.5$^{th}$ |
|---|---|---|---|---|
| Category specific preference: | | | | |
|    Device Tools $\alpha_1$ | -6.22* | 5.04 | -14.327 | -2.669 |
|    eBooks $\alpha_2$ | -11.34* | 3.14 | -15.290 | -6.381 |
|    Games $\alpha_3$ | -4.35* | 3.76 | -11.222 | -2.296 |
|    Health/Diet/Fitness $\alpha_4$ | -4.1 | 2.18 | -5.939 | 2.982 |
|    Humor/Jokes $\alpha_5$ | -16.32* | 5.85 | -22.097 | -9.715 |
|    Internet/WAP $\alpha_6$ | -5.41* | 2.29 | -8.021 | -3.021 |
|    Logic/Puzzle/Trivia $\alpha_7$ | -14.2* | 3.49 | -18.122 | -8.332 |
|    Reference/Dictionaries $\alpha_8$ | -8.48* | 1.92 | -11.092 | -4.547 |
|    Social Networks $\alpha_9$ | -10.54 | 3.47 | -15.530 | 0.076 |
|    University $\alpha_{10}$ | -5.78* | 1.39 | -7.791 | -2.916 |
| States: | | | | |
|    Individual download history State $\alpha_{11}$ | -27.27* | 5.46 | -34.350 | -13.821 |
|    Latent imitation level $\alpha_{12}$ | 0.02* | 0.01 | 0.011 | 0.035 |
| App category characteristics (factors): | | | | |
|    Popularity of app category $\alpha_{13}$ | 1.32 | 0.63 | -0.830 | 1.767 |
|    Investment apps category $\alpha_{14}$ | 5.34 | 1.75 | -0.922 | 7.230 |
|    Hedonic apps category $\alpha_{15}$ | 7.13 | 4.28 | -6.606 | 10.330 |

* $p < 0.05$

**Table 12 PARAMETER ESTIMATES: Individual Choice Hierarchical Model (Local Adopters): CustomerTenure (number of days since registeration on the app-store) explanation of the effects**

| Parameter explained by Tenure | Estimate | Std. Dev. | 2.5th | 97.5th |
|---|---|---|---|---|
| Category specific preference: | | | | |
| Device Tools $\alpha_1$ | -0.00044* | 1.01E-04 | -0.00058 | -0.00023 |
| eBooks $\alpha_2$ | -0.00048* | 2.63E-04 | -0.00087 | -0.00006 |
| Games $\alpha_3$ | -0.00041* | 4.46E-05 | -0.00049 | -0.00032 |
| Health/Diet/Fitness $\alpha_4$ | -0.0008* | 7.30E-05 | -0.00092 | -0.00061 |
| Humor/Jokes $\alpha_5$ | -0.00091* | 2.49E-04 | -0.00126 | -0.00046 |
| Internet/WAP $\alpha_6$ | 0.00011 | 7.58E-05 | -0.00002 | 0.00025 |
| Logic/Puzzle/Trivia $\alpha_7$ | -0.00056* | 1.29E-04 | -0.00081 | -0.00035 |
| Reference/Dictionaries $\alpha_8$ | -0.00028 | 1.50E-04 | -0.00046 | 0.00002 |
| Social Networks $\alpha_9$ | -0.00001 | 9.45E-05 | -0.00016 | 0.00020 |
| University $\alpha_{10}$ | 0.00018 | 1.36E-04 | -0.00007 | 0.00034 |
| States: | | | | |
| Individual download history State $\alpha_{11}$ | -0.00136* | 3.33E-04 | -0.00193 | -0.00081 |
| Latent imitation level $\alpha_{12}$ | 0.00006* | 7.64E-06 | 0.00004 | 0.00007 |
| App category characteristics (factors): | | | | |
| Popularity of app category $\alpha_{13}$ | 0.00001 | 2.06E-05 | -0.00003 | 0.00005 |
| Investment apps category $\alpha_{14}$ | -0.00006 | 6.77E-05 | -0.00016 | 0.00007 |
| Hedonic apps category $\alpha_{15}$ | 0.00021 | 1.35E-04 | -0.00003 | 0.00043 |

\* p<0.05

**Table 13 PARAMETER ESTIMATES: Individual Choice effect**
**(Local Adopters)**

| Total number of users: 1258 | Positive | Negative |
| --- | --- | --- |
| | **Significant** | **Significant** |
| Category specific preference: | | |
| Device Tools $\alpha_1$ | 0 | 1253 |
| eBooks $\alpha_2$ | 0 | 1258 |
| Games $\alpha_3$ | 0 | 1258 |
| Health/Diet/Fitness $\alpha_4$ | 53 | 1205 |
| Humor/Jokes $\alpha_5$ | 0 | 1258 |
| Internet/WAP $\alpha_6$ | 0 | 1258 |
| Logic/Puzzle/Trivia $\alpha_7$ | 0 | 1258 |
| Reference/Dictionaries $\alpha_8$ | 0 | 1258 |
| Social Networks $\alpha_9$ | 53 | 1205 |
| University $\alpha_{10}$ | 0 | 1258 |
| States: | | |
| Individual download history State $\alpha_{11}$ | 0 | 1258 |
| Latent imitation level $\alpha_{12}$ | 1257 | 0 |
| App category characteristics (factors): | | |
| Popularity of app category $\alpha_{13}$ | 1205 | 53 |
| Investment apps category $\alpha_{14}$ | 1205 | 53 |
| Hedonic apps category $\alpha_{15}$ | 1205 | 53 |

**Table 14 COUNTERFACTUAL ANALYSIS: Change in the adoption level by intervening social influence**

| Category specific counterfactual results: | original expected adoption | shut down social influence | 1% more social influence | 1%less social influence | An optimal social influence |
|---|---|---|---|---|---|
| Device Tools | 875.83 | -57% | 0.8% | -0.7% | 55.8% |
| eBooks | 189.45 | -1% | 0.0% | 0.0% | 2.5% |
| Games | 187.51 | 19% | -0.3% | 0.3% | 58.6% |
| Health/Diet/Fitness | 22.21 | 0% | 0.0% | 0.0% | 6.1% |
| Humor/Jokes | 255.09 | 0% | 0.0% | 0.0% | 0.6% |
| Internet/WAP | 1042.20 | 23% | -0.4% | 0.5% | 14.1% |
| Logic/Puzzle/Trivia | 249.12 | 25% | -0.6% | -0.2% | 109.1% |
| Reference/Dictionaries | 1262.09 | 16% | -0.4% | 0.3% | -36.2% |
| Social Networks | 21.66 | 0% | 0.0% | 0.0% | 0.3% |
| University | 18.08 | -1% | 0.0% | 0.0% | 1.5% |
| Total improvement | 4123.25 | 1% | -0.1% | 0.1% | 13.6% |

**Table 15 COUNTERFACTUAL ANALYSIS: Explain optimal social influence improvement with popularity rank of the app category on the app-store**

|  | Coefficients | Standard Error | p-value | t Stat |
|---|---|---|---|---|
| Intercept | -0.185 | 0.147 | 0.245 | -1.254 |
| Category popularity rank | 0.050* | 0.015 | 0.010 | 3.388 |

* p<0.01

**Figure 1 Box and Arrow Representation of the Model**

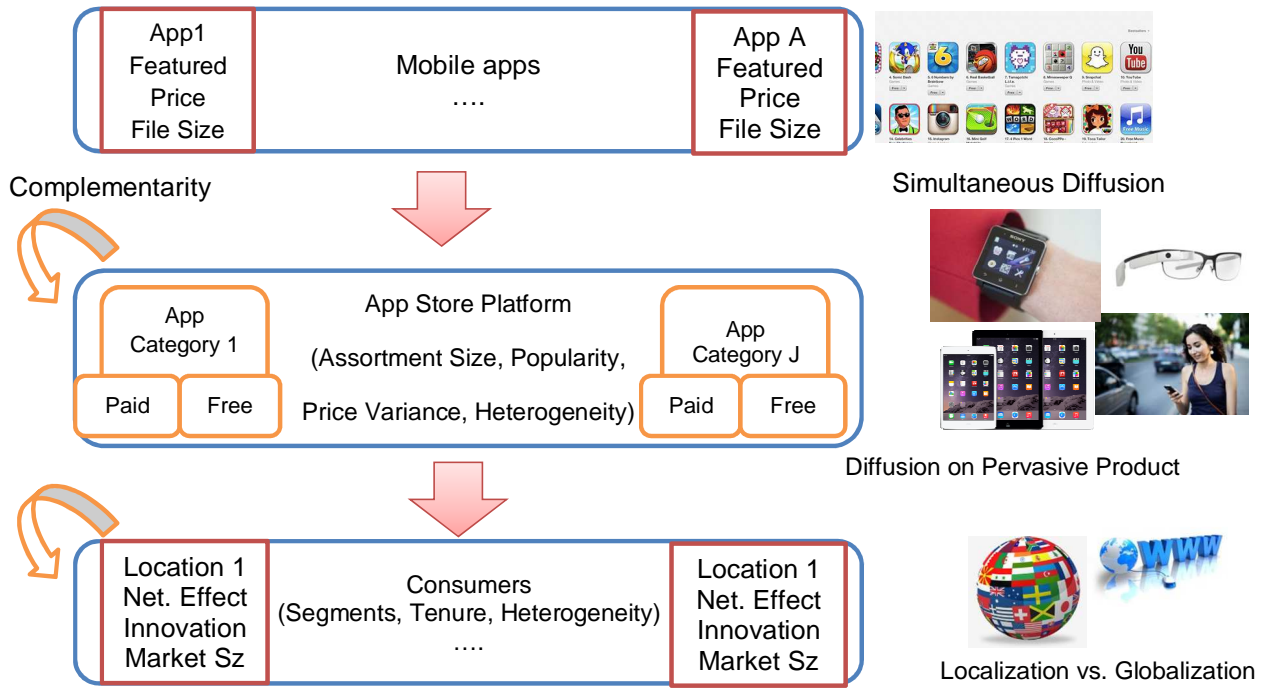

Complementarity

Simultaneous Diffusion

Diffusion on Pervasive Product

Localization vs. Globalization

**Figure 2 Intercontinental (across 30 cities) Diffusion Curves for the mobile apps within the Categories**

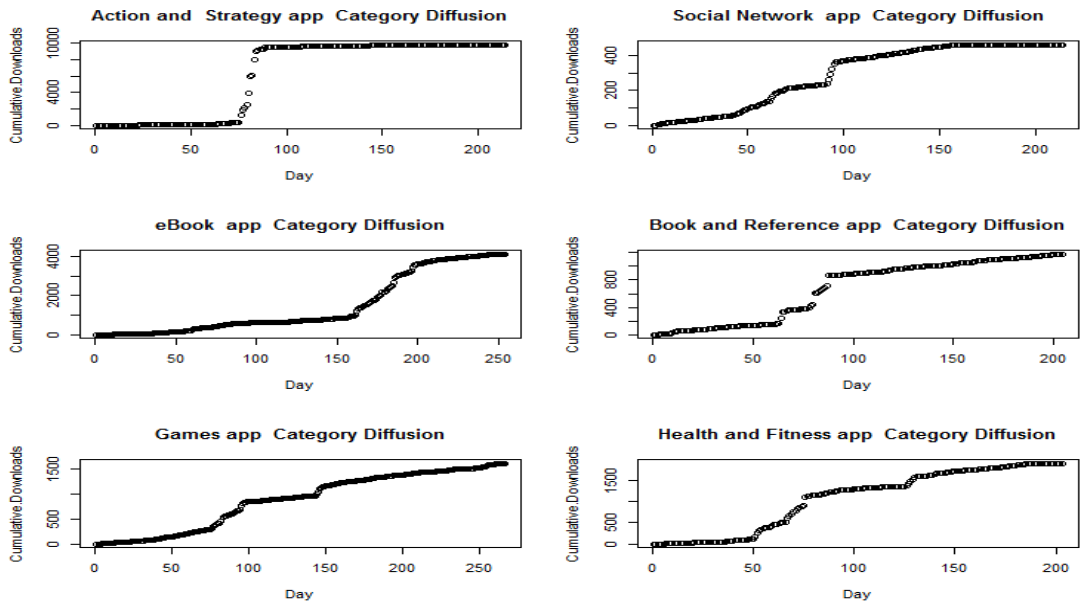

**Figure 3 Popularity (market share) of App Categories on Apple Inc. App Store**

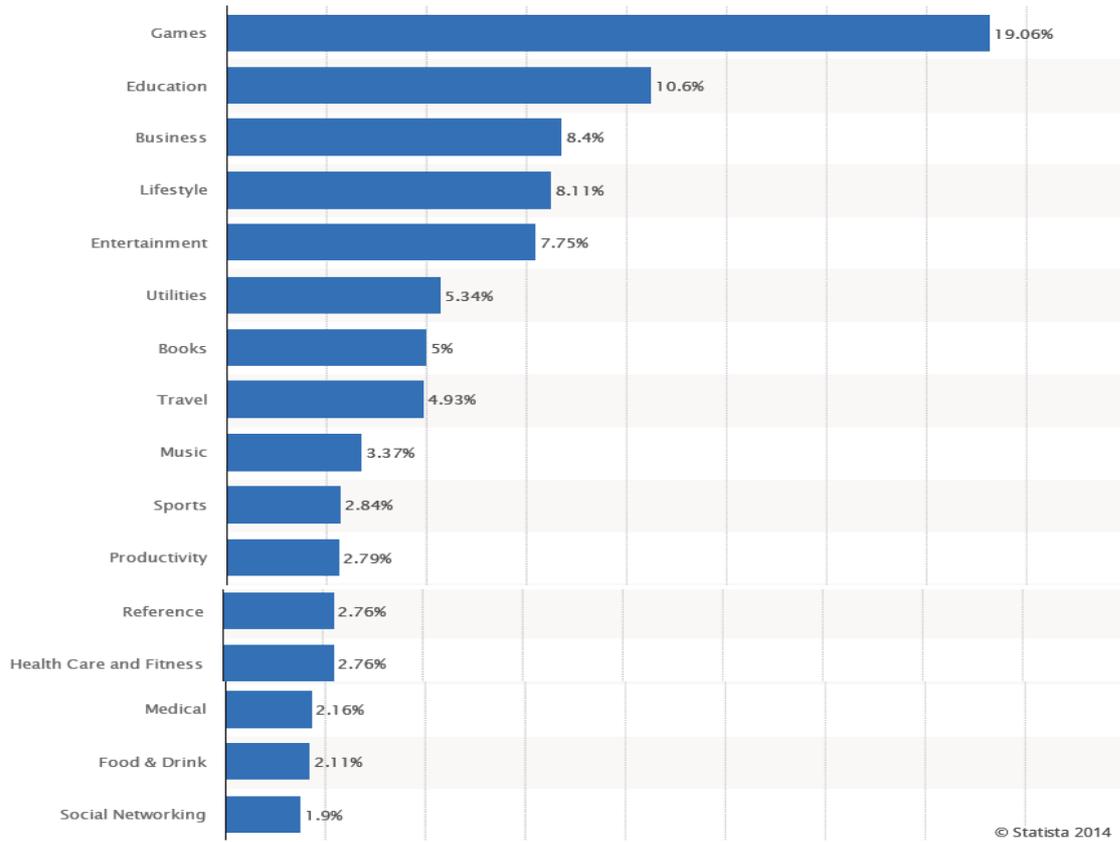

**Figure 4 Free mobile apps versus paid mobile apps**

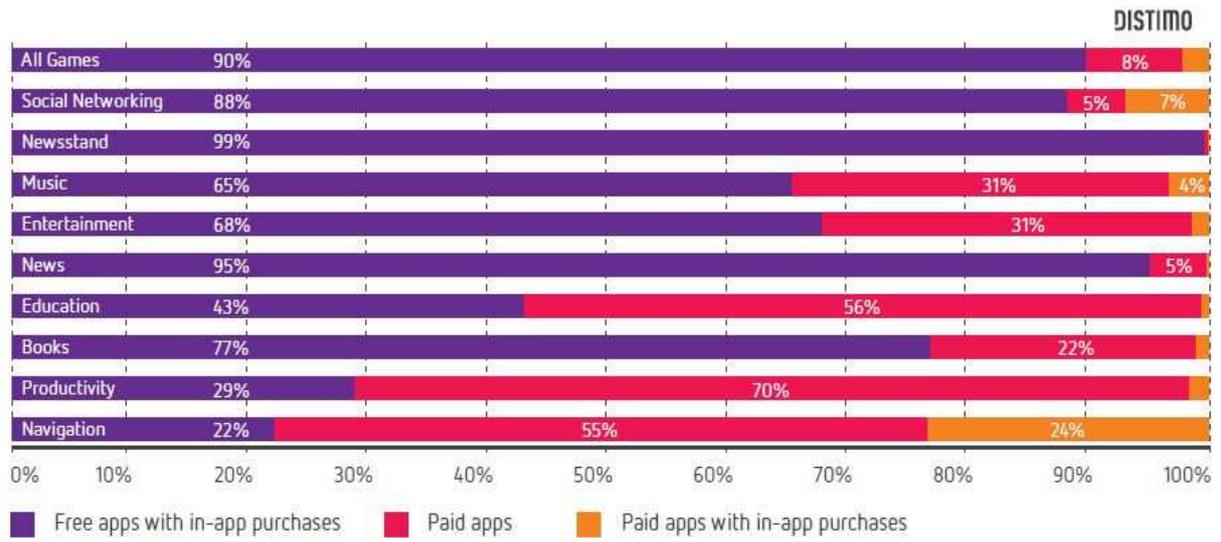

DISTIMO

| | |
|---|---|
| All Games | 90% ... 8% |
| Social Networking | 88% ... 5% 7% |
| Newsstand | 99% |
| Music | 65% ... 31% 4% |
| Entertainment | 68% ... 31% |
| News | 95% ... 5% |
| Education | 43% ... 56% |
| Books | 77% ... 22% |
| Productivity | 29% ... 70% |
| Navigation | 22% ... 55% ... 24% |

0%  10%  20%  30%  40%  50%  60%  70%  80%  90%  100%

■ Free apps with in-app purchases   ■ Paid apps   ■ Paid apps with in-app purchases

**Figure 5 1-Step-ahead Forecast for Local Diffusion (Green Line: a step ahead; Red line: the actual)**

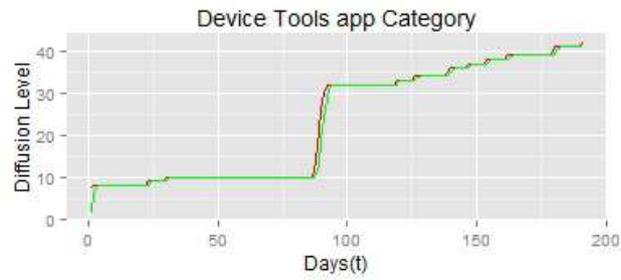

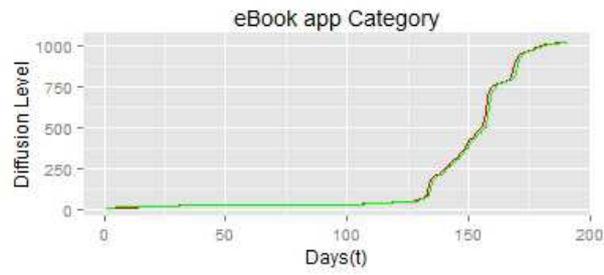

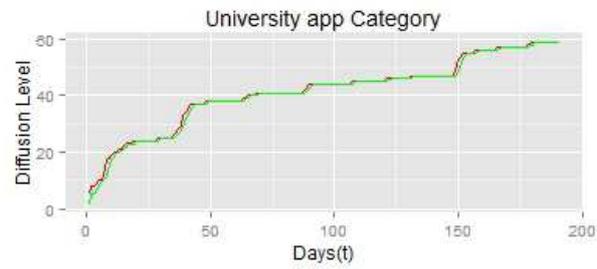

**Figure 6 1-Step-ahead Forecast for Global Diffusion level (Green line: a step ahead, Red line: the actual)**

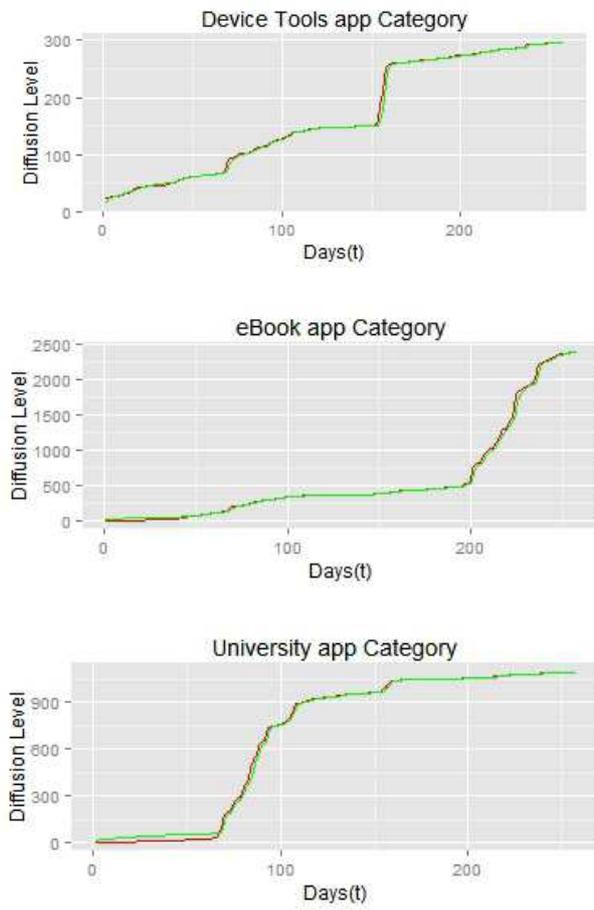

**Figure 7 PARAMETER DISTRIBUTION: Heterogeneity in Individual Choice (Local Adopters)**

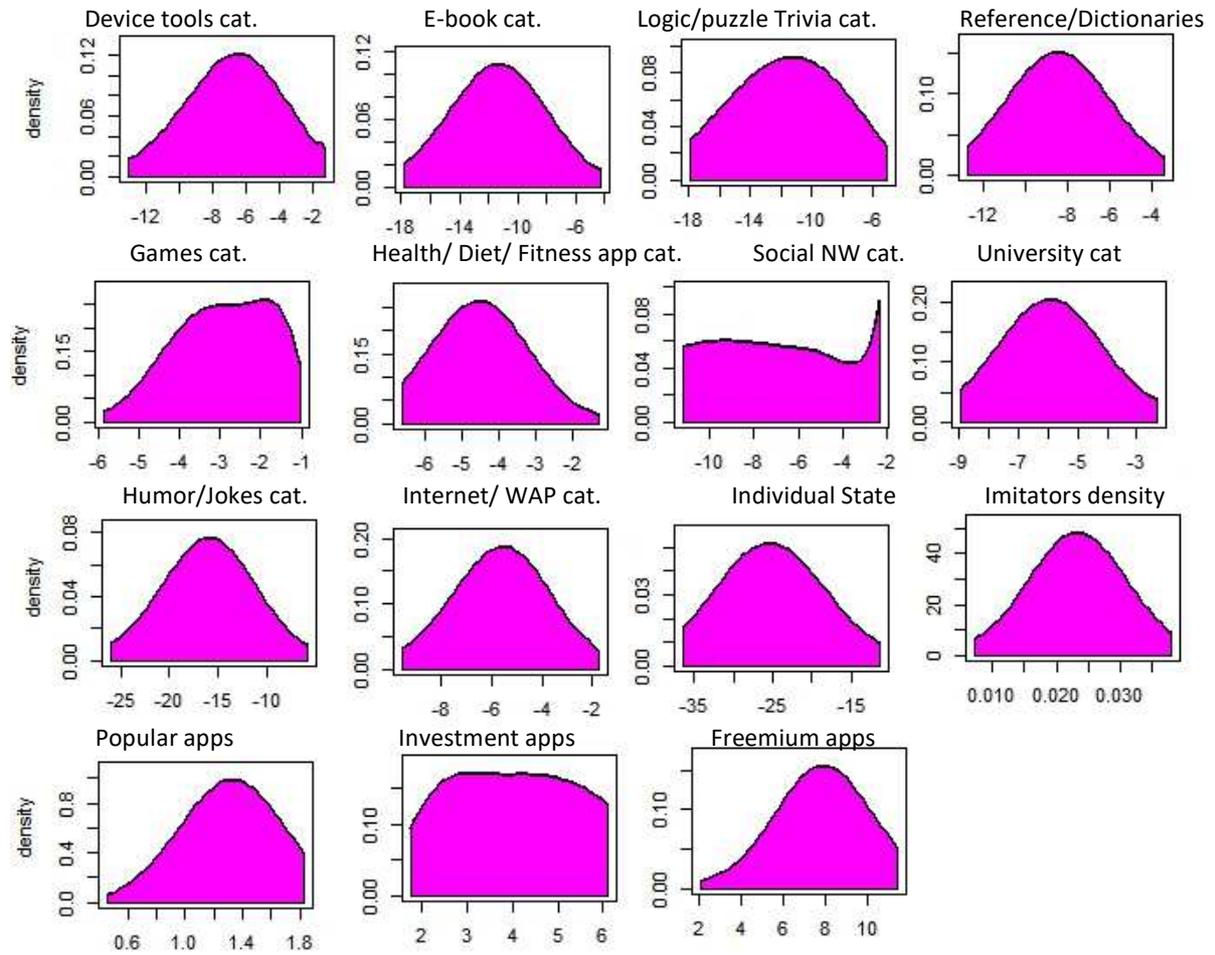

**Figure 8 COUNTERFACTUAL ANALYSIS: an optimal social influence strategy to increase expected adoption level by 14% (log scale)**

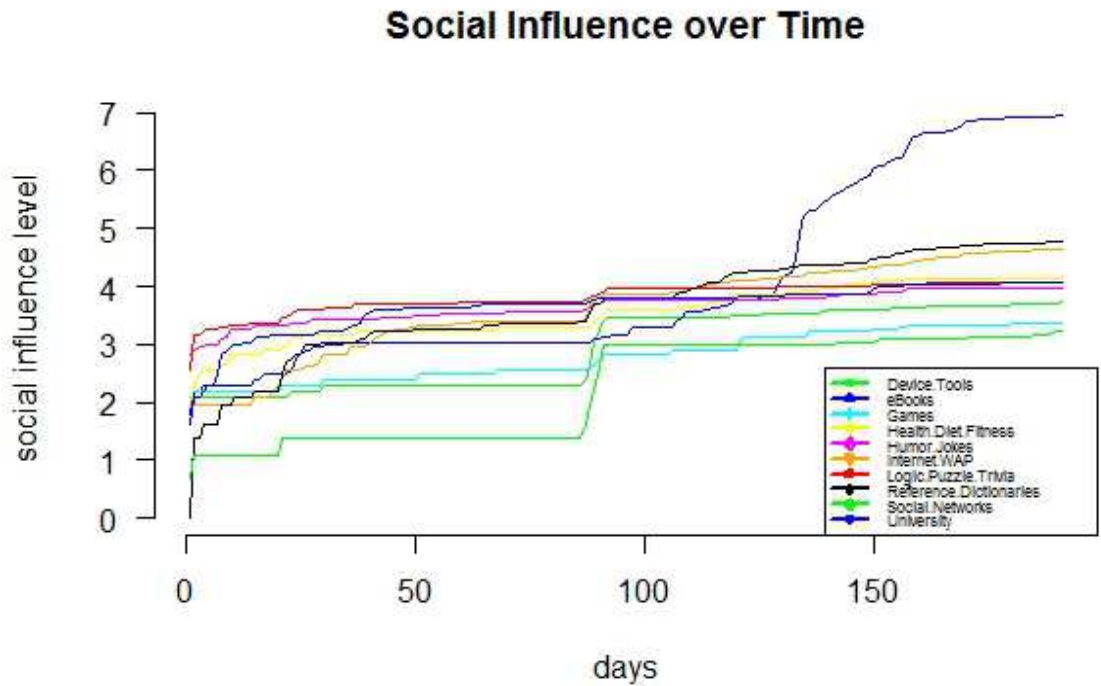

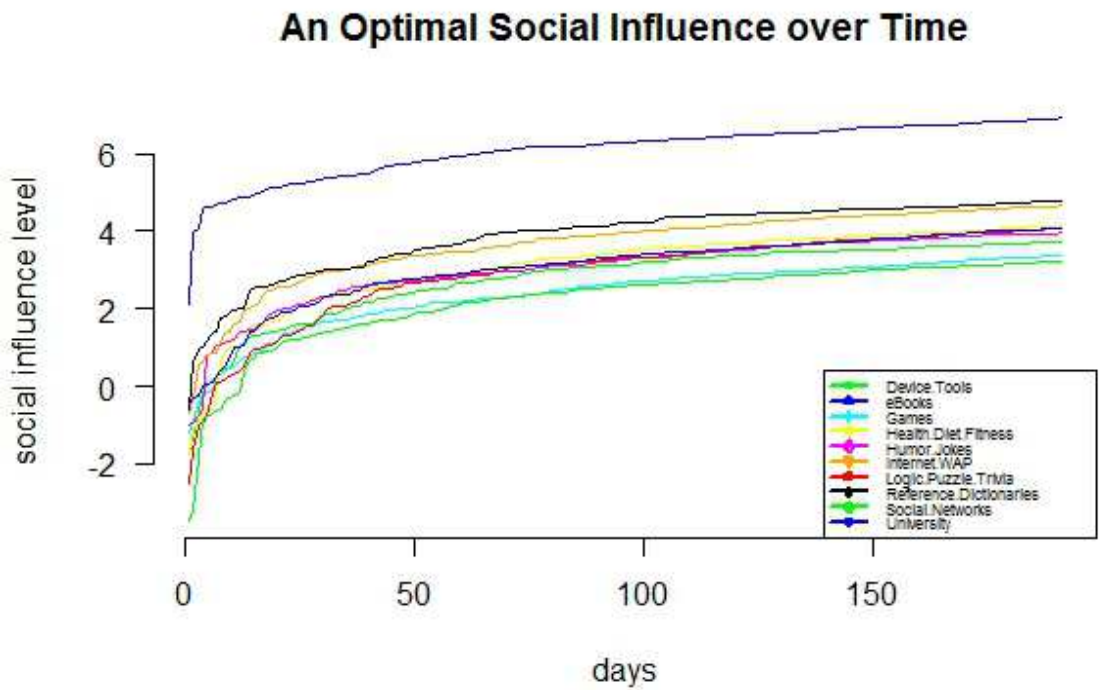

**Social Learning and Diffusion of Pervasive Goods:**

**An Empirical Study of an African App Store**

This Web Appendix contains the following sections:

# APPENDIX A: DIRECT ACYCLIC GRAPH OF CONDITIONAL DISTRIBUTIONS

Probabilistic graphical approaches are popular in computer science, as they not only serve as a visual tool to recognize conditional independence but also help save space in representing probability distributions and facilitate probabilistic queries. Figure A1 represents the probabilistic graphical representation of the model studied in the paper. Shaded circles represent the observed variables, and unshaded circles represent latent variables or parameters. The rectangles, called plates, represent the replication of variables with the number specified at their bottom right.

**Figure A1. Probabilistic graphic of customer mobile app choices under social influence**

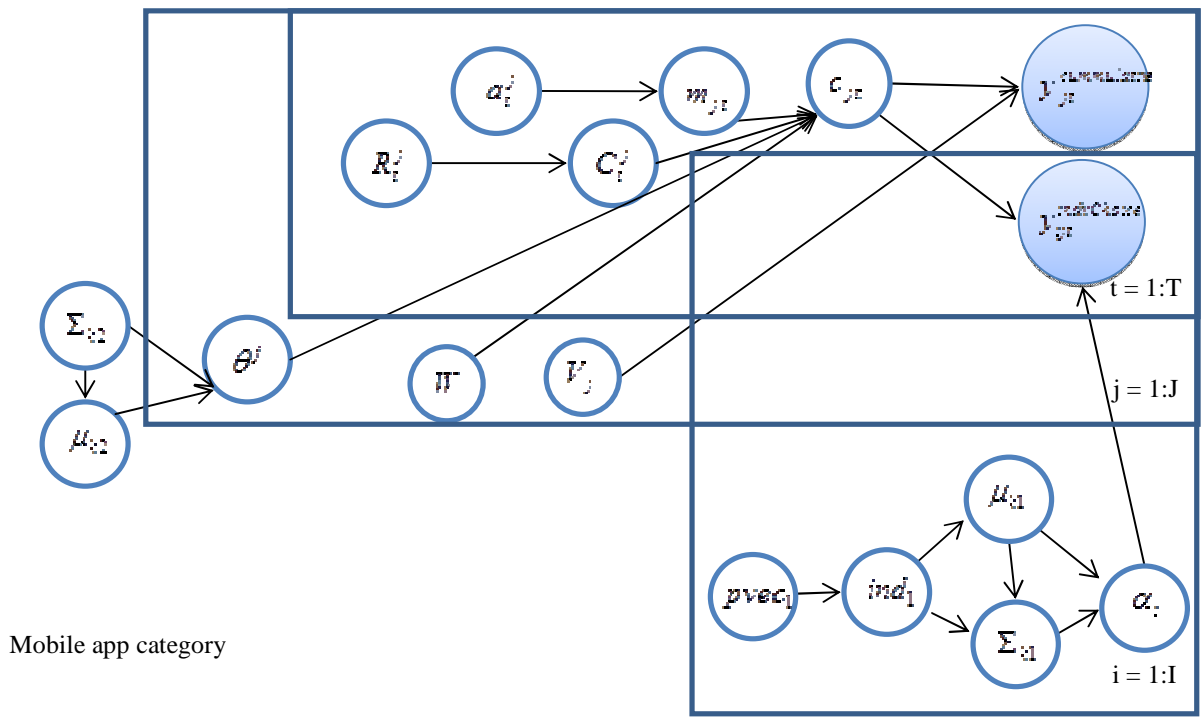

# APPENDIX B: UNSCENTED KALMAN FILTER

A recursive algorithm to update the latent-state variable with the unscented Kalman filter has the following steps (see also Wan and van Der Merwe 2001).

The model has the following form: Equation B1 is an observation equation, and Equation B2 is a state equation, with nonlinear functions H and F:

$$y_k = H(x_k) + \eta_k, \eta_k \sim MVN(0, \mathrm{I})$$
$$x_k = F(x_{k-1}) + \upsilon_k, \upsilon_k \sim MVN(0, \mathrm{Y})$$
(B1)

The estimation algorithm:

$$\widehat{x}_0 = E[x_0]$$
$$P_0 = E[(x_0 - \widehat{x}_0)(x_0 - \widehat{x}_0)']$$
$$k \in \{1, ..., \infty\}$$
$$\lambda = \alpha^2 (L + \kappa) - L$$
$$W_0^m = \frac{\lambda}{L + \lambda}$$
$$W_0^c = \frac{\lambda}{L + \lambda} + (1 - \alpha^2 + \beta)$$
$$W_i^m = W_i^c = \frac{1}{2(L + \lambda)}, i = 1, ..., 2L$$
$$\beta = 2$$

B2)

Drawing sigma points:

$$\aleph_{k-1} = [\widehat{x}_{k-1} \quad \widehat{x}_{k-1} \pm \sqrt{(L + \lambda)P_{k-1}}]$$
(B3)

Updating time:

$$\aleph_{k|k-1} = F[\aleph_{k|k-1}]$$
$$\widehat{x}_k^- = \sum_{i=0}^{2L} W_i^m \aleph_{ik|k-1}$$
$$P_k^- = \sum_{i=0}^{2L} W_i^c [\aleph_{ik|k-1} - \widehat{x}_k^-][\aleph_{ik|k-1} - \widehat{x}_k^-]^T + I$$
(B4)
$$\Im_{k|k-1} = H[\aleph_{k|k-1}]$$
$$\widehat{y}_k^- = \sum_{i=0}^{2L} W_i^m \Im_{ik|k-1}$$

Updating measurement:

$$P_{y_k} = \sum_{i=0}^{2L} W_i^c [\Im_{k|k-1} - \widehat{y}_k^-][\Im_{k|k-1} - \widehat{y}_k^-]^T + Y$$

$$P_{x_k y_k} = \sum_{i=0}^{2L} W_i^c [\aleph_{ik|k-1} - \widehat{x}_k^-][\Im_{k|k-1} - \widehat{y}_k^-]^T$$

$$K = P_{x_k y_k} P_{y_k}^{-1} \qquad\qquad\qquad\qquad (B5)$$

$$\widehat{x}_k = \widehat{x}_k^- + K(y_k - \widehat{y}_k^-)$$

$$P_k = P_k^- - K P_{y_k} K^T$$

## APPENDIX C: CONDITIONAL DISTRIBUTIONS FOR ESTIMATION OF THE MICRO CHOICE MODEL

Conditional distributions of the choice variable include the following:

$$A_i \mid y_{it}^j, s_{it}^j, \mu_i, \Sigma_i \qquad i = 1...I \\ , \hat{c}_{jt}^{imm}, F_{jt} \qquad , \tag{C1}$$

where this conditional distribution can be estimated by the Metropolis–Hasting random walk on the weighted likelihood.

The priors for the normal mixture distribution of the individual- and category-specific parameters used are:

$$\{(\mu_i, \Sigma_i)\} \mid A_i, \Delta, z_i, a, v, \vartheta, \alpha^d$$
$$\alpha^d \mid I *$$
$$a \mid \{(\mu_i, \Sigma_i)*\}$$
$$v \mid \{(\mu_i, \Sigma_i)*\}, \vartheta$$
$$\vartheta \mid \{(\mu_i, \Sigma_i)*\}, v$$

$$\text{C2}$$

where the first conditional is the standard posterior Polya Urn representation for the mean and variance of individual-specific random coefficient choice model parameters and $(\mu_i, \Sigma_i)*$ denotes a set of unique $(\mu_i, \Sigma_i)$, on which the Dirichlet process hyperparameters depend only on (a posteriori). Given the $\{(\mu_i, \Sigma_i)*\}$ set $\alpha^d$ and based measure parameters (i.e. $a, v, \vartheta$) are independent, a posteriori. The conditional posterior of the $G_0$ hyperparameters (i.e., $a, v, \vartheta$) is factored into two parts, as $a$ is independent of $v, \vartheta$ given $\{(\mu_i, \Sigma_i)*\}$. The form of this conditional posterior is:

$$p(a, v, \vartheta \mid \{(\mu_i, \Sigma_i)*\}) \propto \prod_{j=1}^{I^*} \phi(\mu_j^* \mid 0, a^{-1}, \Sigma_j *) IW(\Sigma_j * \mid v, V = v\vartheta I_d) p(a, v, \vartheta) \tag{C3}$$

where $\phi(.|.,.)$ denotes the multivariate normal density and $IW(.|.,.)$ denotes an Inverted-Wishart distribution. Finally, for Polya Urn representation implementation, we use the following conditional distribution:

$$(\mu_{i+1}, \Sigma_{i+1}) \,|\, \{(\mu_1, \Sigma_1),...,(\mu_i, \Sigma_i)\} \sim \begin{cases} G_0(a, v, \vartheta)\ with\ prob & \dfrac{\alpha^d}{\alpha^d + i} \\ \delta_{(\mu_j, \Sigma_j)}\ with\ prob & \dfrac{1}{\alpha^d + i} \end{cases} \quad (C4)$$

We assessed the prior hyperparameters to provide proper but diffuse distributions, defined formally by

$$\vec{a} = 0.00001, \bar{a} = 50, \vec{\vartheta} = 0.00001, \bar{\vartheta} = 600, \vec{v} = 0.00001, \bar{v} = 80 \quad (C5)$$

Finally, to complete the exposition, the posterior for the partition (segment) parameters has the following form:

$$\Sigma_k \,|\, \alpha_k, \Delta, z_k, v, V \sim IW(v + n_k,$$
$$v \times \vartheta \times I + \left(\alpha_k - \tilde{\mu}_k' - \Delta z_k\right)'\left(\alpha_k - \tilde{\mu}_k' - \Delta z_k\right) + a(\tilde{\mu}_k - \bar{\mu})(\tilde{\mu}_k - \bar{\mu}))$$

$$\mu_k \,|\, \alpha_k, \Sigma_k, \bar{\mu}, a, \Delta, z_k \sim N(\tilde{\mu}_k, \frac{\Sigma_k}{n_k + a})$$

$$\tilde{\mu}_k = \frac{n_k \overline{\alpha_k} + a\overline{\mu_k}}{n_k + a}, \overline{\alpha_k} = \frac{\sum_{i \in k} \alpha_i}{n_k}, \overline{\mu} = 0$$

## APPENDIX D: CHOICE PARAMETER ESTIMATES FOR ALTERNATIVE MODELS

Table D1. Parameter Estimates: Individual Choice Effect (Local Imitators)

| | Estimate | Std. Dev. | 2.5th | 97.5th |
|---|---|---|---|---|
| Category specific preference: | | | | |
| Device Tools $\alpha_1$ | -9.19* | 6.57 | -34.490 | -3.216 |
| eBooks $\alpha_2$ | -8.89* | 2.84 | -13.253 | -3.337 |
| Games $\alpha_3$ | -25.82* | 8.31 | -38.609 | -9.519 |
| Health/Diet/Fitness $\alpha_4$ | 1.17 | 1.83 | -6.588 | 2.295 |
| Humor/Jokes $\alpha_5$ | -0.41 | 7.11 | -31.222 | 1.509 |
| Internet/WAP $\alpha_6$ | -12.11* | 3.94 | -18.256 | -4.378 |
| Logic/Puzzle/Trivia $\alpha_7$ | -26.26* | 10.30 | -51.248 | -9.384 |
| Reference/Dictionaries $\alpha_8$ | -17.95* | 5.82 | -27.128 | -6.624 |
| Social Networks $\alpha_9$ | -3.23 | 1.33 | -5.043 | 0.184 |
| University $\alpha_{10}$ | -6.86* | 10.07 | -49.254 | -1.866 |
| States: | | | | |
| Individual download history State $\alpha_{11}$ | -17.79* | 18.37 | -93.057 | -5.689 |
| Latent imitation level $\alpha_{12}$ | 0.02* | 0.01 | 0.005 | 0.032 |
| App category characteristics (factors): | | | | |
| Popularity of app category $\alpha_{13}$ | 0.39 | 0.73 | -2.766 | 0.789 |
| Investment apps category $\alpha_{14}$ | -10.73* | 3.45 | -15.955 | -3.700 |
| Hedonic apps category $\alpha_{15}$ | 12.67 | 7.57 | -15.563 | 20.683 |

* p<0.05

Table D2. Parameter Estimates: Individual Choice Effect (Global Imitators)

| | Estimate | Std. Dev. | 2.5th | 97.5th |
|---|---|---|---|---|
| Category specific preference: | | | | |
| Device Tools $\alpha_1$ | -2.13* | 0.26 | -2.79 | -1.89 |
| eBooks $\alpha_2$ | -0.6 | 0.84 | -0.68 | 0.52 |
| Games $\alpha_3$ | 0.22* | 0.37 | 0.17 | 1.18 |
| Health/Diet/Fitness $\alpha_4$ | 1.06 | 0.86 | -1.46 | 1.53 |
| Humor/Jokes $\alpha_5$ | -2.68* | 0.34 | -3.20 | -1.63 |
| Internet/WAP $\alpha_6$ | -0.53* | 0.28 | -1.76 | -0.44 |
| Logic/Puzzle/Trivia $\alpha_7$ | -1.72 | 0.75 | -2.07 | 1.08 |
| Reference/Dictionaries $\alpha_8$ | -1.45 | 0.46 | -1.76 | 0.64 |
| Social Networks $\alpha_9$ | -1.64* | 0.36 | -1.97 | -0.44 |
| University $\alpha_{10}$ | -2.05* | 0.37 | -2.44 | -1.61 |
| States: | | | | |
| Individual download history State $\alpha_{11}$ | -3.3* | 0.44 | -4.00 | -2.15 |
| Latent imitation level $\alpha_{12}$ | 0.01* | 0.00 | 0.00 | 0.01 |
| App category characteristics (factors): | | | | |
| Popularity of app category $\alpha_{13}$ | -0.08* | 0.15 | -0.38 | -0.06 |
| Investment apps category $\alpha_{14}$ | -0.42* | 1.09 | -1.24 | -0.40 |
| Hedonic apps category $\alpha_{15}$ | -0.31* | 0.49 | -2.29 | -0.22 |

Table D3. PARAMETER ESTIMATES: Individual Choice Effect (Global Adopters)

| | Estimate | Std. Dev. | 2.5th | 97.5th |
|---|---|---|---|---|
| Category specific preference: | | | | |
| Device Tools $\alpha_1$ | -24.94* | 16.81 | -14.327 | -2.669 |
| eBooks $\alpha_2$ | -18.31* | 12.36 | -15.290 | -6.381 |
| Games $\alpha_3$ | -18.66* | 11.97 | -11.222 | -2.296 |
| Health/Diet/Fitness $\alpha_4$ | -13.37 | 8.16 | -5.939 | 2.982 |
| Humor/Jokes $\alpha_5$ | -11.26* | 6.43 | -22.097 | -9.715 |
| Internet/WAP $\alpha_6$ | -6.92* | 2.54 | -8.021 | -3.021 |
| Logic/Puzzle/Trivia $\alpha_7$ | -0.13* | 1.06 | -18.122 | -8.332 |
| Reference/Dictionaries $\alpha_8$ | -17.74* | 12.56 | -11.092 | -4.547 |
| Social Networks $\alpha_9$ | -26.77 | 17.29 | -15.530 | 0.076 |
| University $\alpha_{10}$ | -9.2* | 7.16 | -7.791 | -2.916 |
| States: | | | | |
| Individual download history State $\alpha_{11}$ | -35.93* | 22.88 | -34.350 | -13.821 |
| Latent imitation level $\alpha_{12}$ | 0.01* | 0.01 | 0.011 | 0.035 |
| App category characteristics (factors): | | | | |
| Popularity of app category $\alpha_{13}$ | -1.05 | 0.73 | -0.830 | 1.767 |
| Investment apps category $\alpha_{14}$ | 19.1 | 13.08 | -0.922 | 7.230 |
| Hedonic apps category $\alpha_{15}$ | -25.79 | 17.78 | -6.606 | 10.330 |

Table D4. PARAMETER ESTIMATES: Individual Choice Effect (No Social Influence)

| | Estimate | Std. Dev. | 2.5[th] | 97.5[th] |
|---|---|---|---|---|
| Category specific preference: | | | | |
| Device Tools $\alpha_1$ | -20.51* | 6.14 | -28.162 | -2.250 |
| eBooks $\alpha_2$ | -2.43* | 0.39 | -2.647 | -1.372 |
| Games $\alpha_3$ | -11.89 | 5.26 | -19.040 | 0.482 |
| Health/Diet/Fitness $\alpha_4$ | -10.4* | 3.55 | -15.011 | -0.619 |
| Humor/Jokes $\alpha_5$ | -8.9 | 3.35 | -13.168 | 0.716 |
| Internet/WAP $\alpha_6$ | -21.99* | 7.19 | -31.258 | -2.083 |
| Logic/Puzzle/Trivia $\alpha_7$ | -16.36* | 4.60 | -21.876 | -1.765 |
| Reference/Dictionaries $\alpha_8$ | -7.59 | 1.80 | -8.841 | 0.232 |
| Social Networks $\alpha_9$ | -10.77* | 3.39 | -14.949 | -0.369 |
| University $\alpha_{10}$ | -2.44* | 0.46 | -2.661 | -0.457 |
| States: | | | | |
| Individual download history State $\alpha_{11}$ | -15.84* | 4.79 | -22.241 | -4.298 |
| Latent imitation level $\alpha_{12}$ | - | - | - | - |
| App category characteristics (factors): | | | | |
| Popularity of app category $\alpha_{13}$ | -1.77* | 0.68 | -2.708 | -0.254 |
| Investment apps category $\alpha_{14}$ | -7.51* | 2.81 | -11.425 | -1.362 |
| Hedonic apps category $\alpha_{15}$ | -9.29* | 3.67 | -14.263 | -0.893 |

Table D5. PARAMETER ESTIMATES: Individual Choice Hierarchical Model (Local Imitators): Tenure Explanation of the Effects

| Parameter Explained by Tenure | Estimate | Std. Dev. | 2.5$^{th}$ | 97.5$^{th}$ |
|---|---|---|---|---|
| Category specific preference: | | | | |
| Device Tools $\alpha_1$ | -0.0032* | 9.66E-05 | -0.0034 | -0.0030 |
| eBooks $\alpha_2$ | -0.0012* | 1.42E-04 | -0.0015 | -0.0010 |
| Games $\alpha_3$ | -0.0005* | 1.31E-04 | -0.0008 | -0.0002 |
| Health/Diet/Fitness $\alpha_4$ | -0.0023* | 1.37E-04 | -0.0026 | -0.0021 |
| Humor/Jokes $\alpha_5$ | 0.0006* | 7.44E-05 | 0.0004 | 0.0007 |
| Internet/WAP $\alpha_6$ | 0.0022* | 1.29E-04 | 0.0019 | 0.0023 |
| Logic/Puzzle/Trivia $\alpha_7$ | 0.0028* | 1.60E-04 | 0.0025 | 0.0031 |
| Reference/Dictionaries $\alpha_8$ | 0.0004* | 9.12E-05 | 0.0002 | 0.0006 |
| Social Networks $\alpha_9$ | 0.0034* | 1.46E-04 | 0.0031 | 0.0036 |
| University $\alpha_{10}$ | 0.0007* | 4.04E-05 | 0.0006 | 0.0007 |
| States: | | | | |
| Individual download history State $\alpha_{11}$ | -0.005* | 8.06E-05 | -0.0051 | -0.0048 |
| Latent imitation level $\alpha_{12}$ | 0.0001* | 5.88E-06 | 0.0000 | 0.0001 |
| App category characteristics (factors): | | | | |
| Popularity of app category $\alpha_{13}$ | 0.0001* | 1.13E-05 | 0.0001 | 0.0001 |
| Investment apps category $\alpha_{14}$ | 0.0025* | 6.35E-05 | 0.0024 | 0.0026 |
| Hedonic apps category $\alpha_{15}$ | -0.0012* | 1.09E-04 | -0.0014 | -0.0010 |

* p<0.05

Table D6. PARAMETER ESTIMATES: Individual Choice Hierarchical Model (Global Imitators): Tenure Explanation of the Effects

| Parameter Explained by Tenure | Estimate | Std. Dev. | 2.5th | 97.5th |
|---|---|---|---|---|
| Category specific preference: | | | | |
| Device Tools $\alpha_1$ | -0.0038* | 1.61E-04 | -0.0041 | -0.0035 |
| eBooks $\alpha_2$ | -0.0014* | 1.78E-04 | -0.0017 | -0.0011 |
| Games $\alpha_3$ | 0.0009* | 8.48E-05 | 0.0007 | 0.0010 |
| Health/Diet/Fitness $\alpha_4$ | 0.0046* | 4.85E-04 | 0.0040 | 0.0054 |
| Humor/Jokes $\alpha_5$ | -0.0061* | 3.34E-04 | -0.0065 | -0.0056 |
| Internet/WAP $\alpha_6$ | -0.0005* | 7.52E-05 | -0.0006 | -0.0004 |
| Logic/Puzzle/Trivia $\alpha_7$ | -0.0035* | 1.65E-04 | -0.0038 | -0.0032 |
| Reference/Dictionaries $\alpha_8$ | -0.0033* | 3.75E-04 | -0.0039 | -0.0028 |
| Social Networks $\alpha_9$ | -0.0034* | 2.16E-04 | -0.0037 | -0.0030 |
| University $\alpha_{10}$ | -0.0047* | 2.57E-04 | -0.0051 | -0.0043 |
| States: | | | | |
| Individual download history State $\alpha_{11}$ | -0.0086* | 5.63E-04 | -0.0095 | -0.0077 |
| Latent imitation level $\alpha_{12}$ | -0.0001* | 1.31E-05 | -0.0001 | -0.0001 |
| App category characteristics (factors): | | | | |
| Popularity of app category $\alpha_{13}$ | -0.0002* | 1.72E-05 | -0.0002 | -0.0002 |
| Investment apps category $\alpha_{14}$ | -0.0016* | 9.04E-05 | -0.0018 | -0.0014 |
| Hedonic apps category $\alpha_{15}$ | -0.0005* | 8.27E-05 | -0.0006 | -0.0004 |

* p<0.05

Table D7. PARAMETER ESTIMATES: Individual Choice Hierarchical Model (Global Adopters): Tenure Explanation of the Effects

| Parameter Explained by Tenure | Estimate | Std. Dev. | 2.5$^{th}$ | 97.5$^{th}$ |
|---|---|---|---|---|
| Category specific preference: | | | | |
| Device Tools $\alpha_1$ | 0.0003* | 1.00E-04 | 1.36E-04 | 0.00046 |
| eBooks $\alpha_2$ | 0.00034* | 8.80E-05 | 1.46E-04 | 4.74E-04 |
| Games $\alpha_3$ | -0.00016 | 1.00E-04 | -3.15E-04 | 6.43E-05 |
| Health/Diet/Fitness $\alpha_4$ | 0.00028* | 7.29E-05 | 1.38E-04 | 0.000424 |
| Humor/Jokes $\alpha_5$ | 0.00027* | 9.25E-05 | 1.04E-04 | 0.000454 |
| Internet/WAP $\alpha_6$ | 0.00157* | 1.09E-04 | 1.35E-03 | 1.76E-03 |
| Logic/Puzzle/Trivia $\alpha_7$ | 0.00072* | 9.03E-05 | 5.92E-04 | 0.00091 |
| Reference/Dictionaries $\alpha_8$ | 0.00047* | 6.80E-05 | 3.06E-04 | 5.76E-04 |
| Social Networks $\alpha_9$ | -0.00006 | 1.00E-04 | -2.49E-04 | 9.45E-05 |
| University $\alpha_{10}$ | 0.00071* | 9.72E-05 | 5.37E-04 | 8.64E-04 |
| States: | | | | |
| Individual download history State $\alpha_{11}$ | 0.00119* | 1.98E-04 | 8.84E-04 | 0.001637 |
| Latent imitation level $\alpha_{12}$ | -0.00001 | 4.69E-06 | -1.43E-05 | 3.22E-06 |
| App category characteristics (factors): | | | | |
| Popularity of app category $\alpha_{13}$ | -0.00007* | 1.65E-05 | -9.67E-05 | -3.90E-05 |
| Investment apps category $\alpha_{14}$ | -0.00095* | 1.17E-04 | -1.14E-03 | -7.95E-04 |
| Hedonic apps category $\alpha_{15}$ | 0.0002* | 8.21E-05 | 7.29E-05 | 3.66E-04 |

* $p < 0.05$

Table D8. Parameter Estimates: Individual Choice Hierarchical Model (No Social Influence): Tenure Explanation of the Effects

| Parameter Explained by Tenure | **Estimate** | **Std. Dev.** | **2.5th** | **97.5th** |
|---|---|---|---|---|
| Category specific preference: | | | | |
| Device Tools $\alpha_1$ | 0.00192* | 1.34E-04 | 1.63E-03 | 0.002171 |
| eBooks $\alpha_2$ | 0.00162* | 1.21E-04 | 1.33E-03 | 1.83E-03 |
| Games $\alpha_3$ | 0.00024 | 4.04E-04 | -4.20E-04 | 8.69E-04 |
| Health/Diet/Fitness $\alpha_4$ | -0.00004 | 1.86E-04 | -3.63E-04 | 0.000259 |
| Humor/Jokes $\alpha_5$ | 0.00019 | 1.53E-04 | -6.07E-05 | 0.000484 |
| Internet/WAP $\alpha_6$ | 0.00164* | 2.65E-04 | 1.15E-03 | 2.04E-03 |
| Logic/Puzzle/Trivia $\alpha_7$ | 0.00207* | 1.44E-04 | 1.87E-03 | 0.002432 |
| Reference/Dictionaries $\alpha_8$ | 0.00292* | 1.36E-04 | 2.58E-03 | 3.13E-03 |
| Social Networks $\alpha_9$ | 0.00128* | 1.39E-04 | 1.01E-03 | 0.001511 |
| University $\alpha_{10}$ | 0.00066* | 8.36E-05 | 4.77E-04 | 7.96E-04 |
| States: | | | | |
| Individual download history State $\alpha_{11}$ | 0.00084* | 1.19E-04 | 6.21E-04 | 1.08E-03 |
| Latent imitation level $\alpha_{12}$ | - | - | - | - |
| App category characteristics (factors): | | | | |
| Popularity of app category $\alpha_{13}$ | 0.000005 | 3.77E-05 | -6.62E-05 | 6.02E-05 |
| Investment apps category $\alpha_{14}$ | 0.00023 | 1.65E-04 | -1.11E-04 | 4.58E-04 |
| Hedonic apps category $\alpha_{15}$ | -0.00014 | 2.00E-04 | -4.80E-04 | 1.75E-04 |

* p<0.05

Table D9. Parameter Estimates: Individual Choice Effect
(Local Imitators)

| Total number of users: 1258 | Positive | Negative |
| --- | --- | --- |
| | Significant | Significant |
| Category specific preference: | | |
| Device Tools $\alpha_1$ | 0 | 1258 |
| eBooks $\alpha_2$ | 0 | 1258 |
| Games $\alpha_3$ | 0 | 1258 |
| Health/Diet/Fitness $\alpha_4$ | 1197 | 61 |
| Humor/Jokes $\alpha_5$ | 1197 | 61 |
| Internet/WAP $\alpha_6$ | 0 | 1258 |
| Logic/Puzzle/Trivia $\alpha_7$ | 0 | 1258 |
| Reference/Dictionaries $\alpha_8$ | 0 | 1258 |
| Social Networks $\alpha_9$ | 58 | 1197 |
| University $\alpha_{10}$ | 0 | 1258 |
| States: | | |
| Individual download history State $\alpha_{11}$ | 0 | 1258 |
| Latent imitation level $\alpha_{12}$ | 1217 | 0 |
| App category characteristics (factors): | | |
| Popularity of app category $\alpha_{13}$ | 1197 | 61 |
| Investment apps category $\alpha_{14}$ | 0 | 1258 |
| Hedonic apps category $\alpha_{15}$ | 1197 | 61 |

Table D10. Parameter Estimates: Individual Choice Effect
(Global Imitators)

| Total number of users: 1258 | Positive | Negative |
| --- | --- | --- |
| | **Significant** | **Significant** |
| Category specific preference: | | |
| Device Tools $\alpha_1$ | 0 | 1258 |
| eBooks $\alpha_2$ | 42 | 1216 |
| Games $\alpha_3$ | 1250 | 8 |
| Health/Diet/Fitness $\alpha_4$ | 1208 | 50 |
| Humor/Jokes $\alpha_5$ | 0 | 1257 |
| Internet/WAP $\alpha_6$ | 0 | 1258 |
| Logic/Puzzle/Trivia $\alpha_7$ | 42 | 1216 |
| Reference/Dictionaries $\alpha_8$ | 42 | 1216 |
| Social Networks $\alpha_9$ | 8 | 1250 |
| University $\alpha_{10}$ | 8 | 1250 |
| States: | | |
| Individual download history State $\alpha_{11}$ | 0 | 1258 |
| Latent imitation level $\alpha_{12}$ | 1051 | 4 |
| App category characteristics (factors): | | |
| Popularity of app category $\alpha_{13}$ | 8 | 1250 |
| Investment apps category $\alpha_{14}$ | 8 | 1250 |
| Hedonic apps category $\alpha_{15}$ | 8 | 1250 |

Table D11. Parameter Estimates: Individual Choice Effect
(Global Adopters)

| Total number of users: 1258 | Positive | Negative |
|---|---|---|
| | Significant | Significant |
| Category specific preference: | | |
| Device Tools $\alpha_1$ | 0 | 1258 |
| eBooks $\alpha_2$ | 0 | 1258 |
| Games $\alpha_3$ | 0 | 1258 |
| Health/Diet/Fitness $\alpha_4$ | 0 | 1258 |
| Humor/Jokes $\alpha_5$ | 55 | 1203 |
| Internet/WAP $\alpha_6$ | 0 | 1258 |
| Logic/Puzzle/Trivia $\alpha_7$ | 438 | 545 |
| Reference/Dictionaries $\alpha_8$ | 0 | 1258 |
| Social Networks $\alpha_9$ | 0 | 1258 |
| University $\alpha_{10}$ | 0 | 1249 |
| States: | | |
| Individual download history State $\alpha_{11}$ | 0 | 1258 |
| Latent imitation level $\alpha_{12}$ | 607 | 257 |
| App category characteristics (factors): | | |
| Popularity of app category $\alpha_{13}$ | 55 | 1203 |
| Investment apps category $\alpha_{14}$ | 1203 | 55 |
| Hedonic apps category $\alpha_{15}$ | 55 | 1203 |

Table D12. Parameter Estimates: Individual Choice Effect
(No Social Influence)

| Total number of users: 1258 | Positive | Negative |
|---|---|---|
| | Significant | Significant |
| Category specific preference: | | |
| Device Tools $\alpha_1$ | 0 | 1258 |
| eBooks $\alpha_2$ | 0 | 1258 |
| Games $\alpha_3$ | 54 | 1103 |
| Health/Diet/Fitness $\alpha_4$ | 0 | 1258 |
| Humor/Jokes $\alpha_5$ | 54 | 1204 |
| Internet/WAP $\alpha_6$ | 0 | 1258 |
| Logic/Puzzle/Trivia $\alpha_7$ | 0 | 1258 |
| Reference/Dictionaries $\alpha_8$ | 54 | 1204 |
| Social Networks $\alpha_9$ | 0 | 1258 |
| University $\alpha_{10}$ | 0 | 1258 |
| States: | | |
| Individual download history State $\alpha_{11}$ | 2 | 1256 |
| Latent imitation level $\alpha_{12}$ | - | - |
| App category characteristics (factors): | | |
| Popularity of app category $\alpha_{13}$ | 2 | 1241 |
| Investment apps category $\alpha_{14}$ | 0 | 1243 |
| Hedonic apps category $\alpha_{15}$ | 2 | 1241 |

**Figure D1. Parameter Distribution: Heterogeneity in Individual Choice (Local Imitators)**

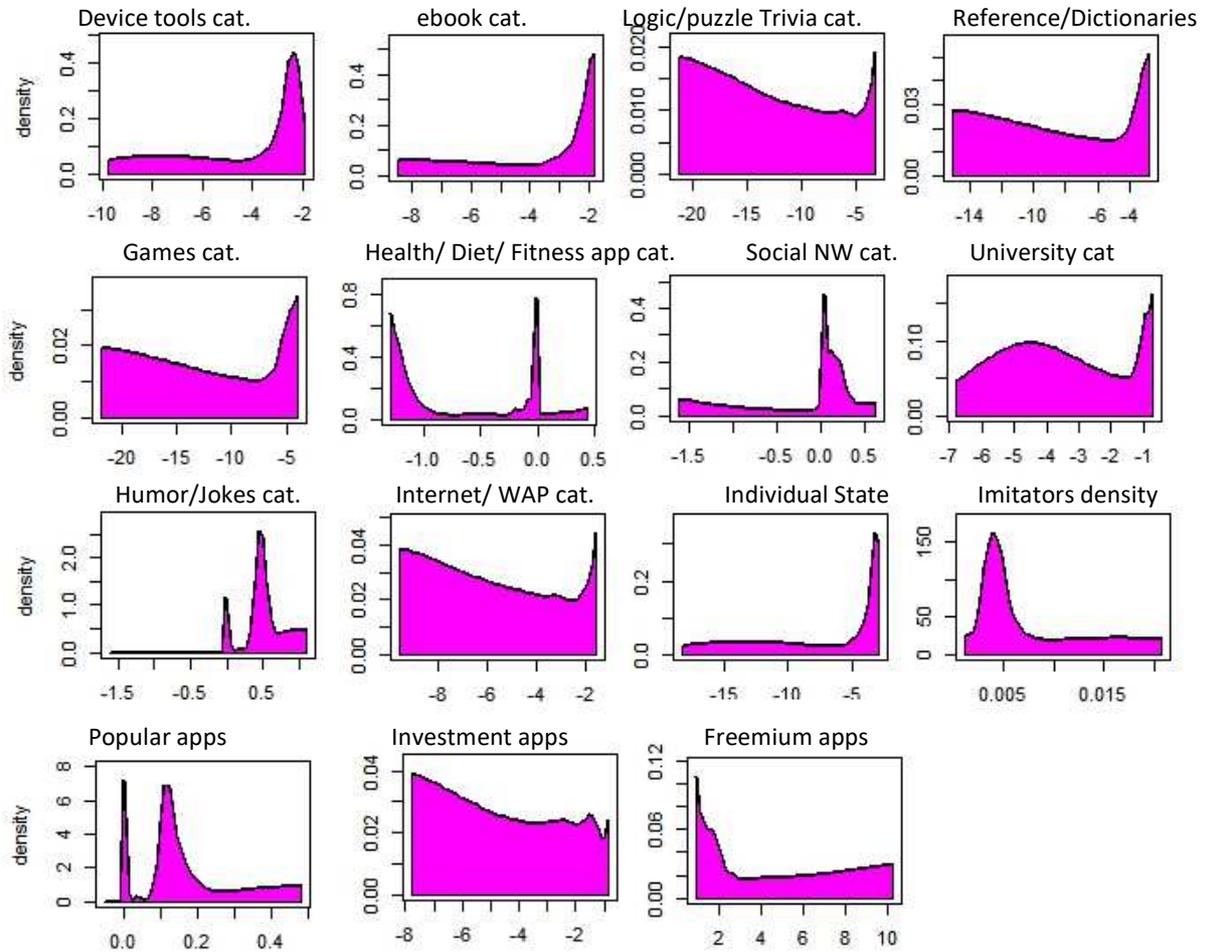

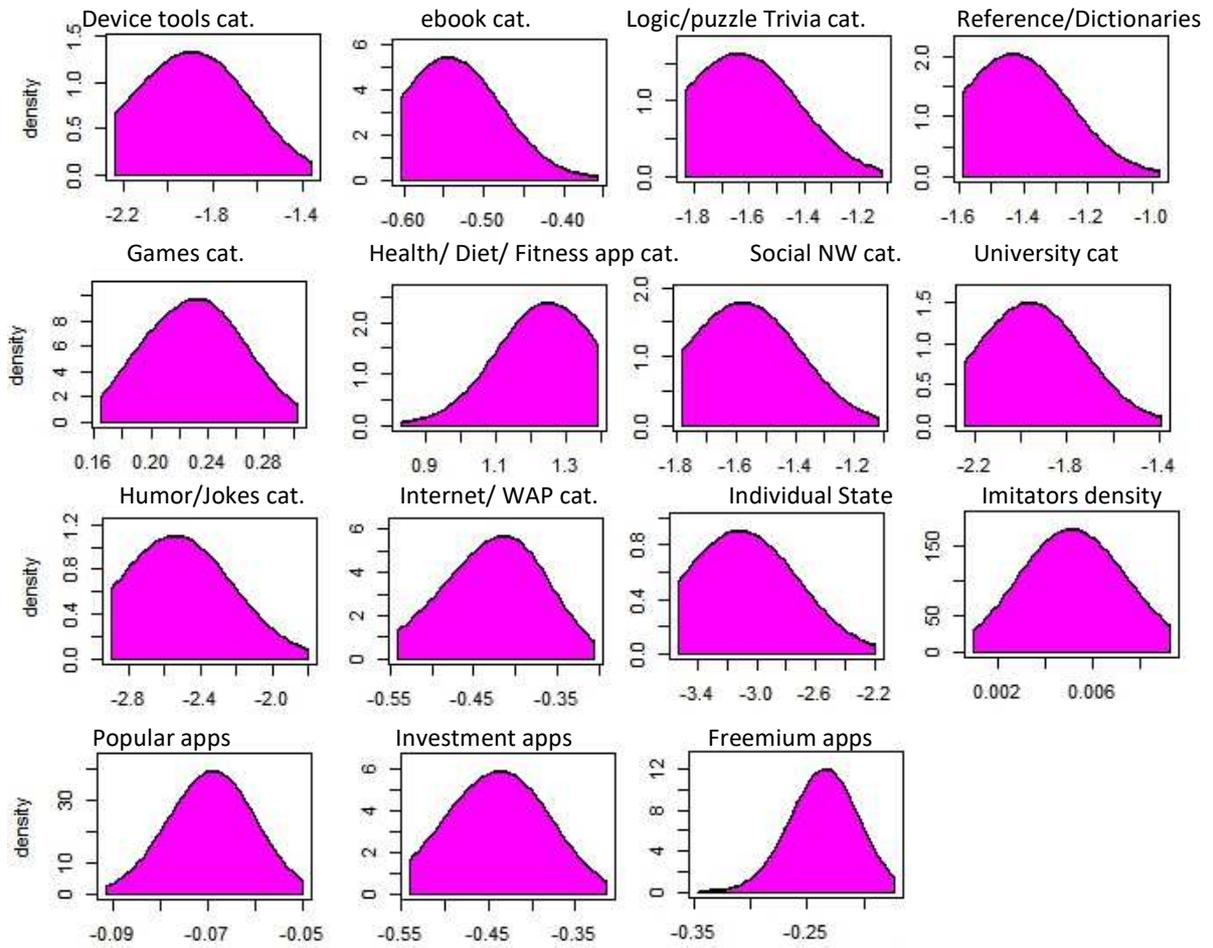

Figure D2. Parameter Distribution: Heterogeneity in Individual Choice (Global Imitators)

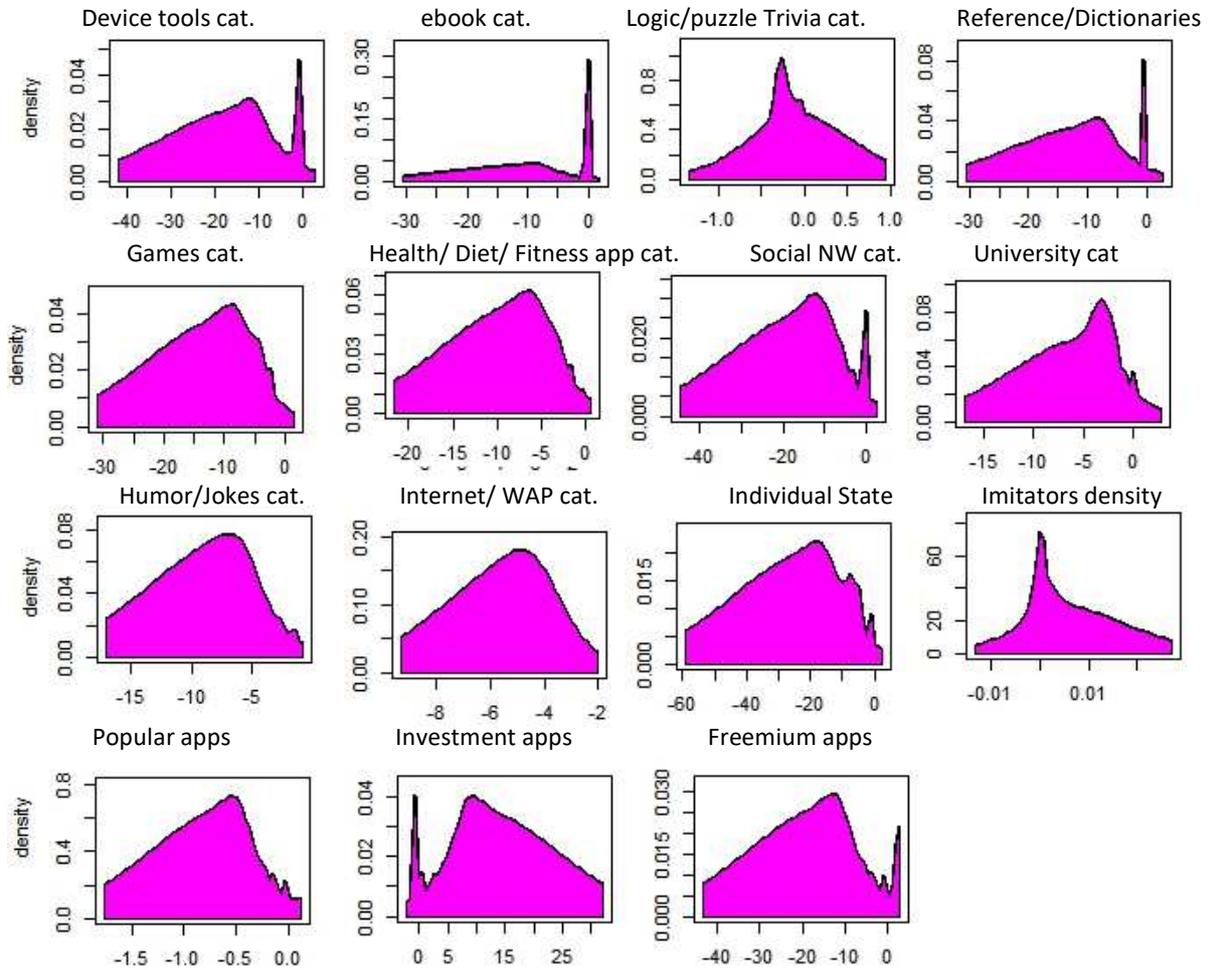

Figure D3. Parameter Distribution: Heterogeneity in Individual Choice (Global Adopters)

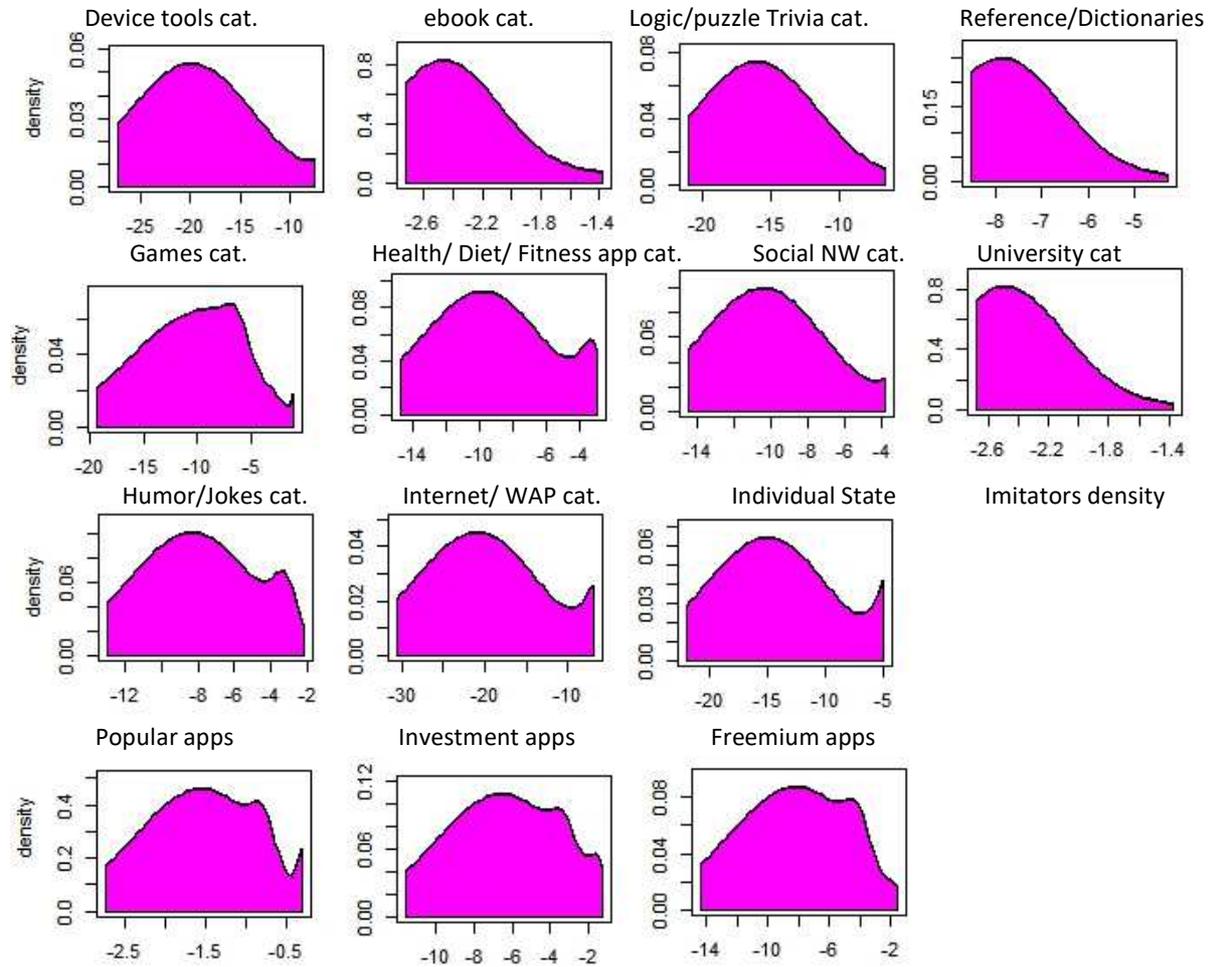

Figure D4. Parameter Distribution: Heterogeneity in Individual Choice (No Social Influence)